\documentclass{article}
\usepackage[utf8]{inputenc}

\pdfoutput=1

\usepackage[backend=biber, style=alphabetic, sorting=nty]{biblatex}
\usepackage{amsmath}
\usepackage{amssymb}
\usepackage{graphicx}
\usepackage{tabularx}
\usepackage[margin=1in]{geometry}
\usepackage[font={small,it}]{caption}
\usepackage{authblk}
\usepackage{hyperref}

\usepackage{placeins}

\let\Oldsection\section
\renewcommand{\section}{\FloatBarrier\Oldsection}

\let\Oldsubsection\subsection
\renewcommand{\subsection}{\FloatBarrier\Oldsubsection}

\let\Oldsubsubsection\subsubsection
\renewcommand{\subsubsection}{\FloatBarrier\Oldsubsubsection}

\newcolumntype{Y}{>{\centering\arraybackslash}X}

\usepackage{xcolor}
\usepackage{soul}

\author[a]{Robert Stephany\thanks{Corresponding Author. 

Email address: \href{mailto:rrs254@cornell.edu}{\texttt{rrs254@cornell.edu}}}$^{,}$}
\author[a,b]{Christopher Earls}
\affil[a]{\it Center for Applied Mathematics, Cornell University, Ithaca, NY 14850, United States}
\affil[b]{\it School of Civil \& Environmental Engineering, Cornell University, Ithaca, NY 14850, United States}
\date{}

\title{PDE-READ: Human-readable Partial Differential Equation Discovery using Deep Learning}

\begin{document}

\maketitle

\begin{center} \textbf{Abstract} \end{center}
PDE discovery shows promise for uncovering predictive models of complex physical systems but has difficulty when measurements are noisy and limited. 
We introduce a new approach for PDE discovery that uses two Rational Neural Networks and a principled sparse regression algorithm to identify the hidden dynamics that govern a system's response. 
The first network learns the system response function, while the second learns a hidden PDE describing the system's evolution. 
We then use a parameter-free sparse regression algorithm to extract a human-readable form of the hidden PDE from the second network.
We implement our approach in an open-source library called \texttt{PDE-READ}.
Our approach successfully identifies the governing PDE in six benchmark examples. 
We demonstrate that our approach is robust to both sparsity and noise and it, therefore, holds promise for application to real-world observational data. 

$~$

\noindent \textbf{Keywords:} Deep Learning, Sparse Regression, Partial Differential Equation Discovery, Physics Informed Machine Learning

\section{Introduction} \label{sec:Intro}

For hundreds of years, humans have sought mechanistic descriptions of the physical world, often in the form of Partial Differential Equations (PDEs).
Historically, scientists discovered these PDEs using a \emph{first-principles} approach; seeking a logical chain of reasoning to explain observed phenomenology.
This approach has yielded myriad scientific and engineering theories, from Newtonian Mechanics to Relativity. 
Notwithstanding the many successes of this approach, progress has been slow on many important fronts: notably power grids, financial markets, protein modeling, drug interaction {\it ect}.
\emph{PDE discovery} is a frontier field that uses machine learning techniques to discover scientific laws, expressed using governing equations, directly from data.
PDE discovery promises to yield predictive models by augmenting the first principles approach with data-driven methods. 

$~$

Neural networks are a natural tool for PDE discovery.
In the landmark paper, Leshno et al. \cite{leshno1993multilayer} showed that for a given non-polynomial activation function, $\sigma$, the set of neural networks with that activation function are dense in the set of compactly supported continuous functions from $\mathbb{R}^n$ to $\mathbb{R}^m$.
Today, this result is known as the \emph{universal approximation property}.
Since the continuous functions with compact support are dense in $L^p$, the set of neural networks must also be dense in $L^p$ \cite{pinkus1999approximation}.
Since both classical and weak form PDE solutions live in $L^p$, the universal approximation property suggests neural networks can approximate the solution of any well-posed PDE. 

$~$

In 2017, \cite{rudy2017data} introduced \texttt{PDE-FIND}: an algorithm that discovers a hidden PDE from a grid of measurements of a system response function, $u$. 
\texttt{PDE-FIND} assumes the time derivative of $u$ can be expressed as a linear combination of \emph{library terms}, $f_1, \ldots, f_{M}$, each of which is a function of $u$ and its spatial partial derivatives. 
Generally, each library term is the product powers of $u$ and its spatial partial derivatives.
\texttt{PDE-FIND} uses finite difference techniques to approximate partial derivatives of $u$ on a suitably specified spatiotemporal grid. 
It then uses these values to evaluate the library terms on the grid.
This engenders a system of linear equations (with one equation for each grid point),

\begin{equation} U_{t} = \Theta(u) \xi, \label{eq:Rudy_System} \end{equation}

where $\Theta(u)$ is a matrix whose $i,j$ element holds the value of the $j$th library term at the $i$th grid point, $U_t$ is a vector whose $i$th element holds the value of the time derivative of $u$ at the $i$th grid point, and $\xi$ is a vector of unknown coefficients.
Finding a $\xi$ that satisfies equation \ref{eq:Rudy_System} reveals the hidden PDE that governs $U$:

\begin{equation} D_{t} u = \xi_1 f_1(u) + \cdots + \xi_M f_M(u) \label{eq:PDE_Rudy}\end{equation}

Where $D_{t} u$ denotes the time derivative of $u$. 
{\texttt{PDE-FIND} determines $\xi$ using STRidge, an iterative algorithm that begins by solving the $L^2$ penalized least-squares problem $\text{argmin}_{\xi}\{\|\Theta(u)\xi - U_{t}\|_2 + \beta \| \xi \|_{2} \}$ (where $\beta$ is a user-defined constant).
This becomes the first candidate solution.
In subsequent steps, given a candidate solution $\xi$, STRidge eliminates every component of $\xi$ whose magnitude is smaller than a user-defined threshold, $\alpha$, and then re-solves the $L^2$ minimized least squares problem using the remaining components.
It repeats this process until the sequence of candidate solutions satisfies a user-specified convergence criterion.}
In their paper, \cite{rudy2017data} showed that \texttt{PDE-FIND} can identify several hidden PDEs --- including Burgers' equation, the Korteweg–De Vries equation, and Schrodinger's equation --- directly from data.

$~$

Despite its foundational importance to PDE discovery, \texttt{PDE-FIND} suffers from a few key limitations: 
First, {\texttt{PDE-FIND} uses finite differences stencils to approximate the derivatives of $u$.
Finite differences can dramatically amplify noise, making it difficult for the algorithm to work with noisy data sets.
Further, these stencils require the data to be on a grid; a potentially cumbersome requirement.}
Second, for the algorithm to work, the user must fine-tune $\alpha$ and $\beta$. 
Adjusting these constants changes the PDE that \texttt{PDE-FIND} uncovers.
Selecting appropriate values can be difficult in practice when the correct equation is not known \emph{a priori}.
{Third,} the user must specify a sufficiently rich library of candidate terms for the algorithm to work.
If there is no way to express the PDE in the form of equation \ref{eq:PDE_Rudy}, then \texttt{PDE-FIND} can not identify the hidden PDE. 
In principle, one can test several libraries until they find one that explains the observed dynamics, but this would necessitate rerunning \texttt{PDE-FIND} several times. 

$~$

Several other PDE discovery algorithms were proposed around the same time as \texttt{PDE-FIND}. 
Two notable examples include \cite{berg2017neural} - which used a neural network to learn the system response function, and then uses sparse regression to identify a PDE that the learned response function satisfies - and \cite{schaeffer2017learning}, whose approach is similar to that of \texttt{PDE-FIND}. 
{Each of these algorithms yielded impressive results, but suffered from similar issues to those described above for \texttt{PDE-FIND}.}
A significant breakthrough came with \cite{both2021deepmod}. 
Their approach, called \texttt{DeepMoD}, learns $\xi$ while simultaneously training a neural network, $U$, to match the measurements of $u$.
Using a network allows the measurements of $u$ to be dispersed anywhere throughout the problem domain.
{Further, doing away with finite differences eliminates their potential to amplify noise.}
\texttt{DeepMoD} uses \emph{Automatic Differentiation} \cite{baydin2018automatic} to calculate the partial derivatives of the neural network, from which it evaluates the library terms. 
Further, since $U$ is defined on the entire problem domain, we can evaluate its derivatives (and, therefore, the library terms) anywhere, not just at the points having observational data.
\texttt{DeepMoD} trains $U$ and $\xi$ using a three part loss function:

$$\mathcal{L}(u, \xi, \theta) = \mathcal{L}_{MSE}(U) + \mathcal{L}_{Reg}(U, \xi) + \mathcal{L}_{L_1}(\xi)$$

$\mathcal{L}_{MSE}$ quantifies how well $U$ matches the measurements of $u$, $\mathcal{L}_{Reg}$ encodes how well $u$ satisfies equation \ref{eq:PDE_Rudy}, and $\mathcal{L}_{L_1}$ is $\lambda$ times the $L_1$ norm of $\xi$.
Here, $\lambda$ is a user-specified hyperparameter.
$\mathcal{L}_{Reg}$ and $\mathcal{L}_{L^1}$ embed the LASSO loss function within \texttt{DeepMoD}'s loss function.
This loss function couples $\xi$ and $U$ such that $\xi$ encodes a PDE which $U$ must satisfy.
$U$ learns to approximate the measurements of $u$, and $\xi$ learns to be sparse, but both must do so while satisfying $U_{t} = \Theta(U)\xi$.
This inductive bias yields a robust algorithm. 
In their paper, \cite{both2021deepmod} demonstrate that \texttt{DeepMoD} can identify PDEs from {noisy and limited data}. 

$~$

Despite this achievement, however, \texttt{DeepMoD} suffers {from two major limitations.}
First, the user must fine-tune $\lambda$. 
Adjusting $\lambda$ effects which PDE the algorithm extracts. 
Thus, \texttt{DeepMoD} only works if the user picks a sage value of $\lambda$, which may be difficult in practical settings.
{Second,} if the user does not select a sufficiently rich library of terms, \texttt{DeepMoD} will not converge. 
Rerunning the algorithm with different libraries is possible but slow, as each run requires retraining $u$ and $\xi$. 
{While some progress has been made in automatically identifying library terms (see, for example, }\cite{khatiry2020evolutionary} {), specifying a sufficiently rich library can still be a practical challenge.}

$~$

Separately, Raissi \cite{raissi2018deep} proposed an innovative approach closely related to PDE discovery: Given a set of noisy measurements of a system response function, $u$, simultaneously train two neural networks, $U$ and $N$. 
$U$ learns an approximation to $u$, while $N$ uses $U$ and its spatial partial derivatives (evaluated using automatic differentiation) to learn the PDE itself. 
To evaluate $N$, we evaluate $U$ and its spatial partial derivatives at a randomly selected set of \emph{collocation points} within the problem domain.
These values are the inputs to $N$, which learns the time derivative of $U$ as a function of $U$ and its spatial partial derivatives.
The associated loss function consists of two parts. 
The first measures how well $U$ conforms to the measurements of the system response function, $u$. 
The second evaluates how well $N$ matches the time derivative of $U$.
This loss function couples $N$ and $U$.
$U$ must learn to match the measurements while also satisfying the learned PDE, $N$.
$U$ cannot simply memorize the measurements and over-fit to their noise. 
The principal deficiency of this approach is that the learned PDE, $N$, lives within a deep neural network and, therefore, can not easily be studied.
This limits the practicality of the approach for discovering new, human-readable PDEs.

$~$

Today, there are many PDE discovery algorithms.
In addition to those methods already discussed, \cite{chen2021physics} (whose method is similar to \texttt{DeepMoD}, but similarly requires the user to tune several hyperparameters and choose a sufficiently rich library before training), \cite{xu2021dl} (whose method is similar to that of \cite{berg2017neural}), \cite{gurevich2019robust} and \cite{messenger2021weak} (who uses a weak-formulation approach), \cite{atkinson2019data} (who uses an evolutionary approach), and \cite{bonneville2021bayesian} (who uses Bayesian Neural networks).
Most of these algorithms involve finding a sparse linear combination of the library of terms that satisfies equation \ref{eq:PDE_Rudy}.
As such, sparse regression is of critical importance to PDE discovery.
Sparse regression seeks a sparse vector $x \in \mathbb{R}^n$ which satisfies $Ax = b$ in the least-squares sense.
Here, $A$ is a matrix and $b \in \mathbb{R}^m$ is a vector.
Ideally, $x$ satisfies

\begin{equation} x = \textrm{argmin} \left\{ \| b - Ax \|_2 + \beta \| x \|_{0} \right\} \label{eq:L0Regression} \end{equation}

For some wisely chosen $\beta > 0$. 
The $L^0$ term encourages sparsity by penalizing the number of non-zero coefficients of $x$. 
There are, however, a few problems with this formulation. 
First, it relies on picking an appropriate $\beta$ value. 
If $\beta$ is too small, $x$ will have too many non-zero components, while if it is too large then $x$ may be too sparse to sufficiently satisfy $Ax = b$. 
Picking an appropriate $\beta$ can be difficult in practice.
Second, solving equation \ref{eq:L0Regression} is NP-hard.
In particular, the only way to find such an $x$ is to solve the least-squares problem $\textrm{argmin}\left\{ \| b - Ax \|_2 : x_k = 0\ \forall \ k \in T \right\}$ for every subset $T \subseteq \{ 1, 2, \ldots, n \}$, and then find which solution satisfies equation \ref{eq:L0Regression}.
Techniques such as STRidge or LASSO effectively weaken the constraints of equation \ref{eq:L0Regression} (namely the $L^0$ term).
These changes can yield sub-optimal solutions, but make the problem computationally feasible.

$~$

Sparse regression is an old problem, however.
Before the birth of PDE discovery, \cite{guyon2002gene} introduced a sparse regression algorithm called \emph{Recursive Feature Elimination} (RFE), to identify genes that predict cancer. 
RFE uses a user-defined \emph{importance metric} to find a sequence of progressively sparser candidate solutions.
The importance metric, roughly speaking, specifies how important a certain component of $x$ is when minimizing $\| Ax - b \|_2$.
The first candidate is the least-squares solution. 
RFE uses an inductive process to find the other candidates:
Given a candidate, $x_k$, RFE identifies the \emph{least important feature} (LIF) of $x_{k}$, and subsequently defines $x_{k+1}$ as the least-squares solution with the LIF removed. 
Thus, RFE derives each successive candidate from the previous one.
We employ RFE in our approach. 

$~$

Existing PDE discovery tools also use conventional activation functions, like $\tanh$ or $\sin$.\footnote{The PDE learning community often avoids $\text{ReLU}$-based networks. A $\text{ReLU}$ network is a piece-wise linear function, which means that its second derivative is zero almost everywhere. Thus, a $\text{ReLU}$ network can not approximate a function's second derivative.}
Recently, however, \cite{boulle2020rational} introduced Rational Neural Networks (RatNNs), whose activation functions are trainable type (3,2) rational functions.
Each layer gets its own trainable rational function (which is applied to each hidden unit in that layer).
Since each activation function contains just $7$ trainable parameters, using RatNNs in place of standard activation functions only slightly increases the number of trainable parameters. 
Importantly, however, the rational activation functions adapt over time, which gives them considerable flexibility over networks with conventional activation functions. 
A RatNN can be viewed as an extremely high order rational function \cite{boulle2020rational}.
This allows RatNNs to take advantage of the approximating power of rational functions.
For example, \cite{boulle2020rational} showed that if $f$ is a function defined on a compact set $S \subseteq \mathbb{R}^n$ and $\varepsilon > 0$, then there exists a Rational Neural Network, $f_{Rat}$, with $\| f - f_{Rat} \|_{\infty} < \varepsilon$ that has far fewer trainable parameters than any given $\text{ReLu}$ network, $f_{ReLu}$, with $\| f - f_{ReLu}\|_{\infty} < \varepsilon$.
This suggests that relatively small Rational Neural Networks can represent complex functions. 
This result makes them ideal candidates for complex scientific problems, like PDE discovery.
This result is not purely theoretical; a recent paper on learning Green's functions, \cite{boulle2022data}, demonstrated that RatNNs learn faster than equivalently sized networks with conventional activation functions.
We employ RatNNs in our approach.

$~$

In this paper, we propose a novel PDE discovery technique that builds upon \cite{raissi2018deep}'s two network approach.
We train a pair of RatNNs, $U$ and $N$, to learn a system response function and a neural network representation of the hidden PDE, respectively. 
Once trained, we use a principled sparse regression technique, based on \cite{guyon2002gene}'s RFE, to extract a human-readable PDE from $N$.
This procedure separates the PDE discovery process into two steps. 
In the first step, we appeal to the universal approximation properties of neural networks to learn the hidden PDE.
In the second step, we attempt to find a sparse linear combination of the library terms which fits $N$.
Notably, our approach is parameter-free.
Further, this approach allows us to quickly test different libraries of candidate terms; we can try fitting each one to $N$ without having to retrain $N$. 
We demonstrate the efficacy of our approach by identifying several PDEs.`

$~$

In section \ref{sub_sec:NN_Approx}, we introduce a two network approach which utilizes RatNNs and periodic re-selection of collocation points.
In section \ref{sub_sec:Regression}, we describe our parameter-free sparse regression algorithm that is based on RFE and is tailored to PDE discovery.
In section \ref{sec:Results}, we demonstrate that our approach can discover linear and non-linear PDEs from {noisy and limited data}.
Finally, in section \ref{sec:Discussion}, we discuss our results, the rationale behind our algorithm, theoretical aspects of our sparse regression technique, and highlight additional aspects of our approach.

\section{Problem Statement and Assumptions} \label{sec:Problem}

Throughout this paper, we will use the following notation to denote partial derivatives: 
Let $f : \mathbb{R}^{n} \to \mathbb{R}$. 
Then $D_{s}^{n} f$ denotes the $n$th partial derivative of $f$ with respect to the variable $s$.
That is, 

$$D_{s}^{n} f = \frac{\partial^n f}{\partial s^n}.$$ 

For brevity, we denote $D_{s}^1 f$ as $D_{s} f$.

$~$

Let $S \subseteq \mathbb{R}$ be an open, connected set, $T > 0$, and $\Omega = (0, T] \times S$.
We assume there exists a \emph{system response function}, $u : \Omega \to \mathbb{R}$, that describes the spatiotemporal evolution of a physical system on $S$ over the time interval $(0, T]$.
For example, $u$ could represent the heat distribution in a structure, or the depth of water within a shallow body. 
We assume we have $N_{Data}$ measurements, $\{ \Tilde{u}(t_i, x_i) \}_{i = 1}^{N_{Data}}$, of $u$ at a collection of points $\{ (t_i, x_i) \}_{i = 1}^{N_{Data}} \subseteq \Omega$. 
We call the points $(t_i, x_i)$ {\it data points}.
We assume the measurements of $u$ are noisy.
Thus, in general, $\Tilde{u}(t_i, x_i) \neq u(t_i, x_i)$, where the latter is the true system response.
We define the {\it noise} at $(t_i, x_i)$ as $u(t_i, x_i) - \tilde{u}(t_i, x_i)$.
We assume the noise is additive at the data points and is i.i.d. Gaussian with mean zero.
We define the level of noise in a collection of noisy measurements as the ratio of the standard deviation of the noise to the standard deviation of the measurements without noise.
We also assume that, in $\Omega$, $u$ satisfies a PDE of the following form:

\begin{equation} D_{t}^{n}u = \hat{N}\Big(u, D_{x} u, D_{x}^{2} u, \ldots , D_{x}^{M} u\Big) \label{eq:PDE}, \end{equation}

$\hat{N}$ is, in general, a nonlinear function of $u$, and its first $M$ spatial partial derivatives. 
In this paper, however, we assume that $\hat{N}$ is a multi-variable polynomial of $u$, $D_{x} u$, \ldots, $D_{x}^{M} u$. In other words, we assume there exists some $K \in \mathbb{N}$ (the degree of the multi-variable polynomial) and constants $c_0, \ldots, c_{M_K} \in \mathbb{R}$ such that

\begin{align*} \hat{N} (u, D_{x} u, \ldots, D_{x}^{M} u ) =\ &c_{0} +  \stepcounter{equation}\tag{\theequation}\label{eq:PolyPDE} \\
&(c_{1})(u) + (c_{2})(D_x u) + \cdots + (c_{M})(D_{x}^{M} u)\ + \\
&(c_{M+1})(u)^2 + (c_{M+2})(u D_{x})(u) + \cdots + (c_{M_2})\left( D_{x}^{M}u \right)^2\ + \\
&\ldots\ + \\
&(c_{M_{K - 1} + 1}) (u)^K + (c_{M_{K - 1} + 2})(u^{K-1})( D_x u) +\cdots + (c_{M_K}) \left( D_{x}^M u \right)^{K}.\end{align*}

\noindent We express this more succinctly as

\begin{equation} \hat{N} \left(u, D_x u, \ldots, D_{x}^{M} u \right) = \left( 1, u, D_x u, \ldots, \left( D_x^{M} u \right)^K \right) \cdot \left(c_0, c_1, \ldots, c_{M_K} \right). \label{eq:PolyPDE:short} \end{equation}

We assume that this representation is sparse, in the sense that most of the $c_k$'s are zero. 
Thus, only a few terms will be present in equation \ref{eq:PolyPDE}.

$~$

For reference, the table below lists the notation in this section.

$~$

\begin{center}
\begin{tabular}{l|l}
    Symbol & Meaning \\
    \hline
    $D_{s}^{n} f$ & The $n$th partial derivative of $f$ with respect to $s$. Same as $\partial^n f / \partial s^n$. \\
    $S$ & Problem spatial domain. A subset of $\mathbb{R}$ \\
    $\Omega$ & The problem domain, $\Omega = (0, T] \times S$, for some $T > 0$. \\
    $u$ & The system response function we are trying to approximate. \\
    $\tilde{u}(t_i, x_i)$ & Noisy measurements of $u$ at $(t_i, x_i) \in \Omega$. \\
    $N_{Data}$ & Number of data points. \\
    $\hat{N}$ & The spatial portion of the PDE that $u$ satisfies (see equation \ref{eq:PDE}). \\
    $M$ & The order of the hidden PDE. \\
    $K$ & Polynomial degree of the representation of $\hat{N}$ \\
\end{tabular}
\end{center}
\section{Methodology} \label{sec:Methodology}

Our goal is to learn the coefficients in equation \ref{eq:PolyPDE:short}.
Our algorithm accomplishes this using a two step procedure, depicted in figures \ref{fig:Method_1} and \ref{fig:Method_2}.
In the first step, we train neural networks $U : \Omega \to \mathbb{R}$ and $N : \mathbb{R}^M \to \mathbb{R}$ to approximate $u$, the system response function, and $\hat{N}$, the right-hand side of equation \ref{eq:PDE}, respectively.
We require that $U$ satisfies the PDE learned by $N$, an approach pioneered by \cite{raissi2018deep}.
Section \ref{sub_sec:NN_Approx} discusses this step in detail. 

$~$

In the second step, we use an adaptation of the Recursive Feature Elimination algorithm \cite{guyon2002gene}, to identify a human-readable PDE from $N$.
In line with our assumptions in section \ref{sec:Problem}, the identified PDE is a multi-variable polynomial of $u$ and its spatial partial derivatives. 
Section \ref{sub_sec:Regression} describes this step in detail.

\subsection{Neural Network Approximations to $u$ and $\hat{N}$} \label{sub_sec:NN_Approx}

\begin{figure}[ht]
    \centering
    \includegraphics[width=\linewidth]{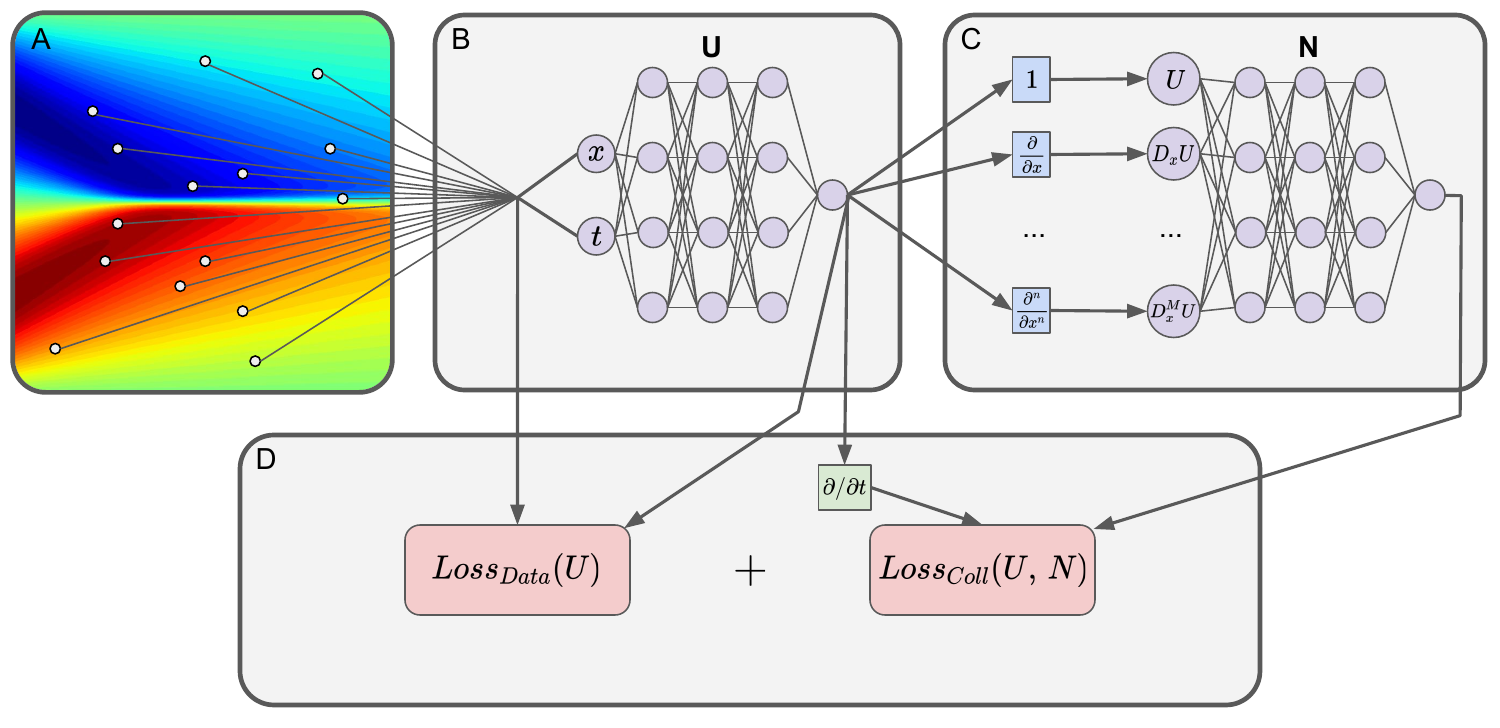}
    \caption{This figure depicts the first step of our algorithm. We break this step into four sub-steps, shown in boxes A-D. In step A, we measure the system response function (whose plot is the background of box A). The circles in this box are the data points. In step B, we evaluate $U$ at each data point. In step C, we evaluate $U$ and its spatial partial derivatives at the collocation points, which are uniformly distributed throughout the problem domain. We use these values to evaluate $N$. In step D, we evaluate the loss functions. The value of the system response and $U$ at the data points determine $Loss_{Data}$, while the time derivative of $U$, and the outputs of $N$, determine $Loss_{Coll}$. We then use back-propagation to update both networks' parameters based on this loss. This procedure repeats each epoch.}
    \label{fig:Method_1}
\end{figure}

We define two RatNNs, $U$ and $N$. 
$U$ learns to approximate the system response function, $u$, using the {noisy and limited} measurements $\{ \Tilde{u}(t_i, x_i) \}_{i = 1}^{N_{Data}}$, while $N$ learns an approximation to the hidden PDE, $\hat{N}$. 
Since we can not enforce the hidden PDE on all of $\Omega$ (which is uncountably infinite), we instead enforce it on a random collection of points in $\Omega$. 
In particular, we draw $N_{Coll}$ samples from a uniform distribution over $\Omega$.
We then train $U$ to satisfy the learned PDE at the resulting set of {\it collocation points}, $\{ (t'_i, x'_i) \}_{i = 1}^{N_{Coll}} \subseteq \Omega$.
Here, $N_{Coll} \in \mathbb{N}$ is a user-specified hyperparameter.

$~$

We re-select the collocation points every epoch\footnote{A training Epoch is the set of computations required for the networks to train on the entire set of training data, once.} by drawing $N_{coll}$ new samples from a uniform distribution over $\Omega$.
Re-selecting the collocation points effectively increases the number of points at which we enforce the hidden PDE, which helps $N$ learn the hidden PDE. 
We discuss this point in section $\ref{sub_sec:Method_Discuss}$.

$~$

We evaluate $U$ at each data point, $(t_i, x_i)$, and compare these values to the measurements, $\{ \tilde{u}(t_i, x_i) \}_{i = 1}^{N_{Data}}$.
Next, we evaluate $U$, its first time derivative, and its first $M$ spatial partial derivatives at each collocation point. 
We use these values to evaluate $N$ at the collocation points.
This allows us to evaluate the {\it PDE residual},

\begin{equation} R_{PDE}(t, x) = \left| D_{t}^{n}U(t, x) - N\left( U(t, x), D_{x}U(t, x), \ldots, D_{x}^M U(t, x) \right)\right|. \label{eq:Residual:PDE} \end{equation}

We then use the PDE residual, the measurements of $\hat u$, and the values of $U$ at the data points, to construct the following loss function:

\begin{equation} Loss(U, N) = \frac{1}{N_{Data}}\sum_{i = 1}^{N_{Data}} \left| U(t_i, x_i) - \Tilde{u}(t_i, x_i) \right|^2 + \frac{1}{N_{Coll}} \sum_{j = 1}^{N_{Coll}} R_{PDE}(t'_j, x'_j)^2 \label{eq:Loss} \end{equation}

For future reference, we define the first term on the right-hand side of equation \ref{eq:Loss} as the {\it data loss}, denoted $Loss_{Data}$, and the second as the {\it collocation loss}, denoted $Loss_{Coll}$.
Thus,

$$Loss(U, N) = Loss_{Data}(U) + Loss_{Coll}(U, N).$$

Our approach thus far mirrors the one established by \cite{raissi2018deep}, except that we re-sample the collocation points and utilize RatNNs. 
Each layer in $N$ and $U$ uses a trainable rational activation function (a trainable type of $(3, 2)$ rational function, which we apply to each neuron in that layer). 
The coefficients in these activation functions train alongside the network weights and biases.
We initialize our rational activation functions using the coefficients proposed in \cite{boulle2020rational}, which give the best type $(3, 2)$ rational approximation to a ReLU. 

$~$

We train $N$ and $U$ using a combination of the \texttt{Adam} \cite{kingma2014adam} and \texttt{LBFGS} optimizers \cite{liu1989limited}. 
We begin training using the \texttt{Adam} optimizer with a learning rate of $0.001$, and then adjust the learning rate depending on how quickly the networks converge. 
In particular, if convergence stops and the value of the loss begins to oscillate, then we decrease the learning rate. 
On the other hand, if the networks converge slowly (with the loss steadily decreasing), then we increase the learning rate. 
Further, in many cases, we switch to the \texttt{LBFGS} optimizer once the \texttt{Adam} optimizer partially trains the networks. 
We train both networks until the loss stops decreasing for either $200$ epochs of the \texttt{Adam} optimizer, or $10$ epochs of the \texttt{LBFGS} optimizer.
As such, each experiment requires a different number of epochs, which we report with our experimental results.

$~$

Figure \ref{fig:Method_1} depicts an overview of this step. 
The table below summarizes the notation in this section.

$~$
\begin{center}
\begin{tabular}{l|l}
    Symbol & Meaning \\
    \hline
    $U$ & Neural Network to approximate $u$ \\
    $N$ & Neural Network to approximate $\hat{N}$ \\
    $\{ (t_i', x_i') \}_{i = 1}^{N_{Coll}}$ & Collocation points \\
    $N_{Coll}$ & Number of collocation points \\
    $R_{PDE}(t, x)$ & PDE residual at the collocation point $(t, x)$ \\
    $Loss(U, N)$ & The loss function. 
\end{tabular}
\end{center}

$~$

\subsection{Sparse Regression via Recursive Feature Elimination} \label{sub_sec:Regression}

\begin{figure}[ht]
    \centering
    \includegraphics[width=\linewidth]{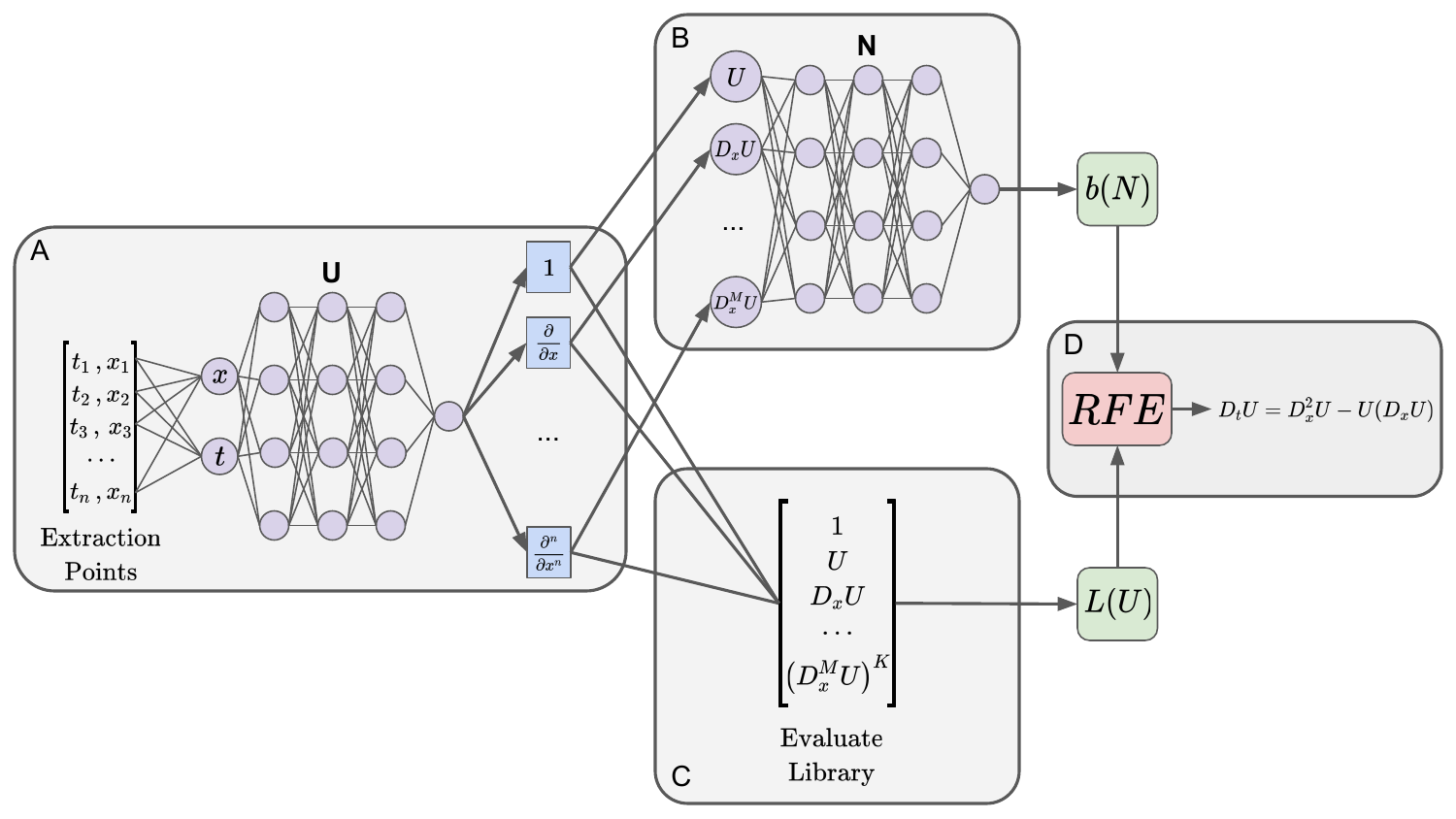}
    \caption{This figure depicts the second step of our algorithm. We break this step into four sub-steps, shown in boxes A-D. Both networks are trained before this step. In step A, we generate the extraction points and evaluate $U$ and its spatial partial derivatives at these points. In step B, we use these values to evaluate $N$ at the extraction points. We use these values to construct the vector $b(N)$, whose $i$th component holds the value of $N$ at the $i$th extraction point. In step C, we use the value of $U$ and its spatial partial derivatives to evaluate the library of candidate terms at the extraction points. We use these values to construct $L(U)$, whose $i,j$ entry holds the value of the $j$th library term at the $i$th extraction point. In step D, we Recursive Feature Elimination (RFE) to find a sequence of sparse candidate solutions to $b(N) \approx L(U)c$. RFE then ranks these candidate solutions and reports the highest-ranking candidates.}
    \label{fig:Method_2}
\end{figure}

After training $U$ and $N$, we identify a human-readable PDE from $N$. 
Our approach is inspired by that of \cite{rudy2017data} and \cite{schaeffer2017learning}. 
Recall that we assume $\hat{N}$ satisfies equation $\ref{eq:PolyPDE:short}$, which we restate here for convenience: 

\begin{equation} \hat{N} \left(U, D_x U, \ldots, D_{x}^{M} U \right) = \left( 1, U, D_x U, \ldots, \left( D_x^{M} U \right)^K \right) \cdot \left(c_0, c_1, \ldots, c_{M_K} \right) \label{eq:PolyPDE:Short:2} \end{equation}

Once $N$ trains, we expect it to approximately satisfy equation \ref{eq:PolyPDE:Short:2}.

$~$

We call the functions $1, U, D_{x} U, \ldots, (D_{x}^M U)^K$ on the right-hand side of \ref{eq:PolyPDE:Short:2} the \emph {library terms}.
To determine the coefficients $c_0, c_1, \ldots, c_{M_K}$, we draw $N_{Extract}$ samples from a uniform distribution over $\Omega$.
We call the resulting coordinates {\it extraction points}. 
Here, $N_{Extract}$ is another user-specified hyperparameter. 
At each extraction point, $(t, x)$, we evaluate $U$, along with its first $M$ spatial partial derivatives. 
Using these values, we calculate each term occurring on the right-hand side of equation \ref{eq:PolyPDE:Short:2}.
We subsequently construct the {\it Library of candidate terms}, $L(U)$, at these locations.
$L(U)$ is an $N_{Extract}$ by $M_K$ matrix, whose $i,j$ entry holds the value of the $j$th candidate term at the $i$th extraction point.
Let $b(N) \in \mathbb{R}^{N_{Extract}}$ be the vector whose $i$th component holds the value of $N$ at the $i$th extraction point. 
Finally, let $c = (c_0, c_1, \ldots, c_{M_K}) \in \mathbb{R}^{M_K}$.
Then,

\begin{equation} b(N) \approx L(U)c. \label{eq:LeastSquares} \end{equation}

We seek a sparse vector, $c$, which satisfies this equation in the least-square sense.
Finding the coefficient vector, $c$, therefore boils down to a sparse regression problem. 
To find $c$, we propose a new method that builds on Guyon et al.'s Recursive Feature Elimination (REF) method \cite{guyon2002gene}. 
First, we normalize $L(U)$ such that each of its columns has unit $L^2$ norm. 
To do this, we define $\tilde{L}(U)$ as follows:

$$\tilde{L}(U)_k = \frac{L(U)_k}{\| L(U)_k \|_2}$$

Where, in general, $A_k$ denotes the $k$th column of some arbitrary matrix $A$.
With this, equation \ref{eq:LeastSquares} becomes the following:

\begin{equation} b(N) \approx \tilde{L}(U) c' \label{eq:ModLeastSquares} \end{equation} 

Where $c'_k = c_k\| L(U)_k \|_2$. 
We apply RFE to this linear system, and base our feature importance metric on the least-squares residual of a candidate solution.
If $c$ is a candidate, then the least-squares residual of $c$, denoted $R_{LS}(c)$, is given by

$$R_{LS}(c) = \| L(U)c - b(N)\|_2^2 = \| \tilde{L}(U)c' - b(N) \|_2^2.$$

We define a feature’s {\it importance} as the amount the residual increases if we set that feature to zero, while holding the other features fixed.\footnote{
Note that the least important feature is not necessarily the feature, $c_k$, that would give the smallest residual if we re-solve the least-squares problem subject to the constraint $x_k = 0$.
We do, however, expect these two features to be the same in most cases. 
We discuss this point further in section \ref{sub_sub_sec:Sparse_Discuss}.}
Mathematically, the $k$th feature's importance is 

$$R_{LS}(c - c_ke_k) - R_{LS}(c).$$

In the discussion section, we show that because the columns of $\tilde{L}(U)$ have unit $L^2$ norm, the least important feature in $c$ is the one whose corresponding component of $c'$ has the smallest magnitude. 
{Normalizing the columns of $L(U)$ also normalizes the library terms according to their magnitude within the problem domain. 
For example, the library terms $U$ and $U^5$ will, in many cases, have very different magnitudes. 
If the magnitude of one term is significantly larger than the others, then the least-squares solution to equation} \ref{eq:LeastSquares} {may be very sensitive to the component of $c$ corresponding to that term. 
Normalizing eliminates this risk.} 

$~$

RFE yields a sequence of increasingly sparse candidate solutions.
Let $c^k$ denote the candidate solution from the $k$th step of this procedure (equivalently, the $k$th ``least sparse solution.'') 
Since $c^{k+1}$ solves $b \approx L(U)c^{k}$ subject to the constraint that the least important feature of $c^{k}$ is removed, we must have $R_{LS}(c^{k+1}) \geq R_{LS}(c^k)$.
Thus,

$$ R_{LS}(c^1) \leq R_{LS}(c^2) \leq \ldots \leq R_{LS}(c^{M_K}) \leq R_{LS}(0) = \| b \|_2^2.$$

With this in mind, for each $k$, we evaluate the ranking metric,

\begin{equation} \frac{R_{LS}(c^{k+1})}{R_{LS}(c^k)}. \label{eq:Ranking_Metric} \end{equation}

Note that when $k = M_K$, we set $c^{k + 1} = 0$.\footnote{This is consistent with the case when $k \neq M_K$, since the $M_K$th candidate will have just one feature, meaning that if we remove its least important feature, then we get the $0$ vector.}
The ranking metric of a candidate effectively tells us how much the PDE residual increases if we remove the LIF from that candidate. 
We rank the candidates according to the values given by equation \ref{eq:Ranking_Metric}. 
In principle, the highest-ranking candidates contain only features necessary to describe the hidden dynamics. 
Critically, this approach uses no tunable parameters. 
Our algorithm reports the top five candidate solutions as the human-readable PDEs.
It also reports the least-squares residual and the ranking metric (reported as a percent) for each of these top five candidates. 

$~$

Figure \ref{fig:Method_2} depicts an overview of this step. 
The table below summarizes the notation in this section.

$~$
\begin{center}
\begin{tabularx}{\textwidth}{l|X}
    Symbol & Meaning \\
    \hline
    $N_{Extract}$ & The number of extraction points \\
    $b(N)$ & An $N_{Extract}$ element vector whose $i$th entry contains the value of $N$ evaluate at the $i$th extraction point. \\
    $L(U)$ & A matrix whose $i,j$ entry holds the value of the $j$th library term at the $i$th extraction point. \\
    $\tilde{L}(U)$ & A matrix whose $k$th column satisfies $\tilde{L}(U)_k = L(U)_k/\| L(U)_k \|_2$ \\
    $R_{LS}(c)$ & The least-squares residual $\| L(U)c - b(N)\|_{2}^2 = \| \tilde{L}(U)c' - b(N) \|$ 
\end{tabularx}
\end{center}
\section{Results} \label{sec:Results}

We implement the approach detailed in section \ref{sec:Methodology} as an open-source Python library.
We call our implementation {\it PDE Robust Extraction Algorithm for Data}, or {\it \texttt{PDE-READ}} for short. 
\texttt{PDE-READ} uses PyTorch and its autodiff capabilities (to evaluate the derivatives of $U$) \cite{paszke2017automatic}.
At the time of writing, no Neural Network framework has native support for rational activation functions; thus, we implement them explicitly in PyTorch.
Our entire library is available at \url{https://github.com/punkduckable/PDE-READ}.

$~$

In this section, we test \texttt{PDE-READ} on six benchmark examples: 
\begin{enumerate}
    \item The Heat equation (Section \ref{sub_sec:Heat})
    \item Burgers' equation (Section \ref{sub_sec:Burgers})
    \item The Allen-Cahn equation (Section \ref{sub_sec:Allen-Cahn})
    \item The Kortwewg-De Vries (KdV) equation (Section \ref{sub_sec:KdV})
    \item The Klein-Gordon (KG) equation (Section \ref{sub_sec:KG})
    \item The dynamic beam equation (Section \ref{sub_sec:Beam})
\end{enumerate}
For each equation, we generate one or more data sets using the Chebfun \cite{driscoll2014chebfun} package in \texttt{MATLAB}.\footnote{We use the \texttt{expm} class to generate the heat equation data sets, the \texttt{Chebop2} class to generate the KG and dynamic beam equation data sets, and the \texttt{spin} class to generate the Burgers, Allen-Cahn, and KdV equation data sets.}
The \texttt{MATLAB} sub-directory of our available library includes the scripts that we use to generate the data sets.

$~$

These experiments demonstrate that \texttt{PDE-READ} can discover a wide array of linear and non-linear equations, often in the presence of extreme noise and sparsity.
In addition to demonstrating our algorithm's efficacy, these experiments also reveal interesting properties and key limitations of the \texttt{PDE-READ} algorithm.
Our experiments with the heat equation in section \ref{sub_sec:Heat} reveal a non-uniqueness issue that is fundamental to PDE-Discovery. 
Our experiments with the Korteweg–De Vries equation in section \ref{sub_sec:KdV} demonstrate our sparse regression algorithm works well even when the library of candidate terms is quite large.
Our experiments with the Klein-Gordon and dynamic beam equations in sections \ref{sub_sec:KG} and \ref{sub_sec:Beam}, respectively, demonstrate that \texttt{PDE-READ} can learn PDEs that involve higher-order time derivatives.

$~$

\noindent We use the following procedure to generate a {noisy and limited} data set with $N_{Data}$ data points and $p$\% noise:

\begin{enumerate}
    \item Calculate the standard deviation, $\sigma_{nf}$, of the noise-free data set. 
    \item For each data point in the noise-free data set, sample a Gaussian distribution with mean $0$ and standard deviation $(p/100)* \sigma_{nf}$.
    Add these values to the noise-free data set to obtain a noisy data set. 
    \item Randomly select an $N_{Data}$ element subset of the noisy data set (by sampling a uniform distribution over the collection of all subsets of data points with $N_{Data}$ elements). 
\end{enumerate}
The resulting collection comprises our {noisy and limited} data set.

$~$

Further, unless stated otherwise, we use the following architecture in these experiments:

\begin{itemize}
\item $U$ and $N$ are RatNNs.
\item $U$ has $5$ hidden layers with $50$ hidden units per layer. 
\item $N$ has $2$ hidden layers with $100$ hidden units per layer. 
\end{itemize}

Furthermore, after training, we identify a PDE using $20,000$ extraction points ($N_{Extract} = 20,000$).

$~$

Many of our experiments include a plot to visualize \texttt{PDE-READ}'s solution.
Each plot depicts the learned solution, the noisy data set (in its entirety, not just the subset we train on), and two measures of error in the trained networks.
The first, the {\it absolute error}, is the absolute value of the difference between the learned solution $U$ and the noise-free data set. 
This metric visualizes how well $U$ reconstructs the noise-free data set from the {noisy and limited} one.
The second is the PDE residual (defined by equation \ref{eq:Residual:PDE} in section \ref{sub_sec:NN_Approx}).
This metric helps us understand how well $U$ satisfies the learned PDE on the problem domain, $\Omega$.

$~$

In these experiments, the {\it true PDE} is the one we use to generate the data set, while the {\it identified PDE} is the highest-ranking candidate PDE that \texttt{PDE-READ} uncovers.
Our principal goal is for the identified PDE to contain the same terms as the true PDE. 
Our secondary coal is for the coefficients in the identified PDE to match the corresponding coefficients in the true PDE.
Critically, we can only satisfy the second goal if we satisfy the first.
Assuming the identified and true PDEs have the same terms, we measure our second goal using the relative error between the coefficients in the identified PDE and the corresponding coefficients in the true PDE. 
In particular, if a coefficient's value in the true PDE is $c_{t}$, and the coefficient's value in the identified PDE is $c_{e}$, then the relative error between them is 

$$\left|\frac{c_{e} - c_{t}}{c_{t}}\right|.$$ 

$~$

\subsection{Heat Equation} \label{sub_sec:Heat}

\begin{figure}[!hbt]
    \centering
    \includegraphics[width=.6\linewidth]{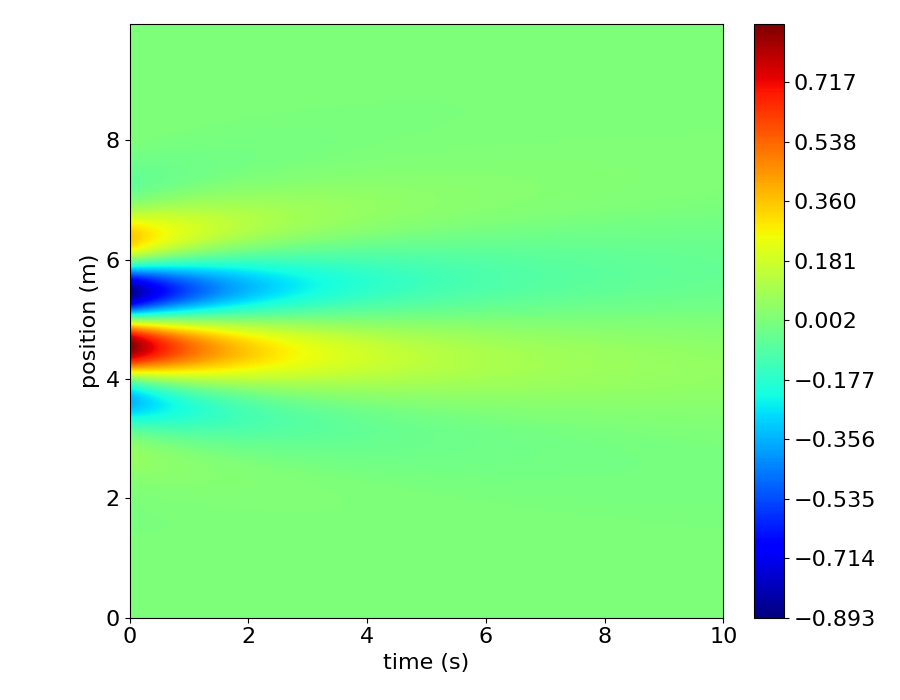}
    \caption{First Heat equation data set (noise-free).}
    \label{fig:Heat_Sine_Exp_Dataset}
\end{figure}

The heat equation describes how heat diffuses through a body over time. 
In one dimension, the heat equation with constant thermal diffusivity $\alpha$ is

\begin{equation} D_t U = -\alpha(D_x^2 U). \label{eq:Heat} \end{equation}

Here, $U(t, x)$ represents the temperature at time $t$ and position $x$.
We test two data sets for this equation, both with $\alpha = 0.05$.

$~$

For these data sets, we use periodic boundary conditions and set $\Omega = (0, T] \times S = (0, 10] \times (0,10)$.
To generate the data sets, we partition each of $S = (0, 10)$ and $(0, T] = (0, 10]$ into $200$ equally-sized sub-intervals (each with a width of $0.05$).
This engenders a uniform grid with $201$ equally-spaced grid lines along both axes.
We then use Chebfun's \texttt{expm} class to solve the heat equation on this grid.
Both data sets contain $40,401 = 201^2$ data points. 

$~$

For each experiment with the Heat equation, our library of candidate PDE terms is 

$$\left\{ \left(U\right)^{p_0}\left(D_x U\right)^{p_1}\left(D_x^2 U\right)^{p_2}: p_0, p_1, p_2 \in \mathbb{N}_{0} \text{ and } p_0 + p_1 + p_2 \leq 2 \right\}.$$

\noindent {\bf First data set:} For the first heat equation data set, we use Chebfun to solve the heat equation on $\Omega$ with the following initial condition:

$$U(x, t) = \exp \left( -0.5(x - 5)^2 \right) \sin \left( 5x \frac{2\pi}{5} \right) \quad\quad x \in (0, 10).$$

Figure \ref{fig:Heat_Sine_Exp_Dataset} depicts the noise-free data set.
For this experiment, the {noisy and limited} data set contains $10,000$ data points ($N_{Data} = 10,000$) and $100$\% noise. 
We also use $10,000$ collocation points ($N_{Coll} = 10,000$).

\begin{figure}[!hbt]
    \centering
    \includegraphics[width=\linewidth]{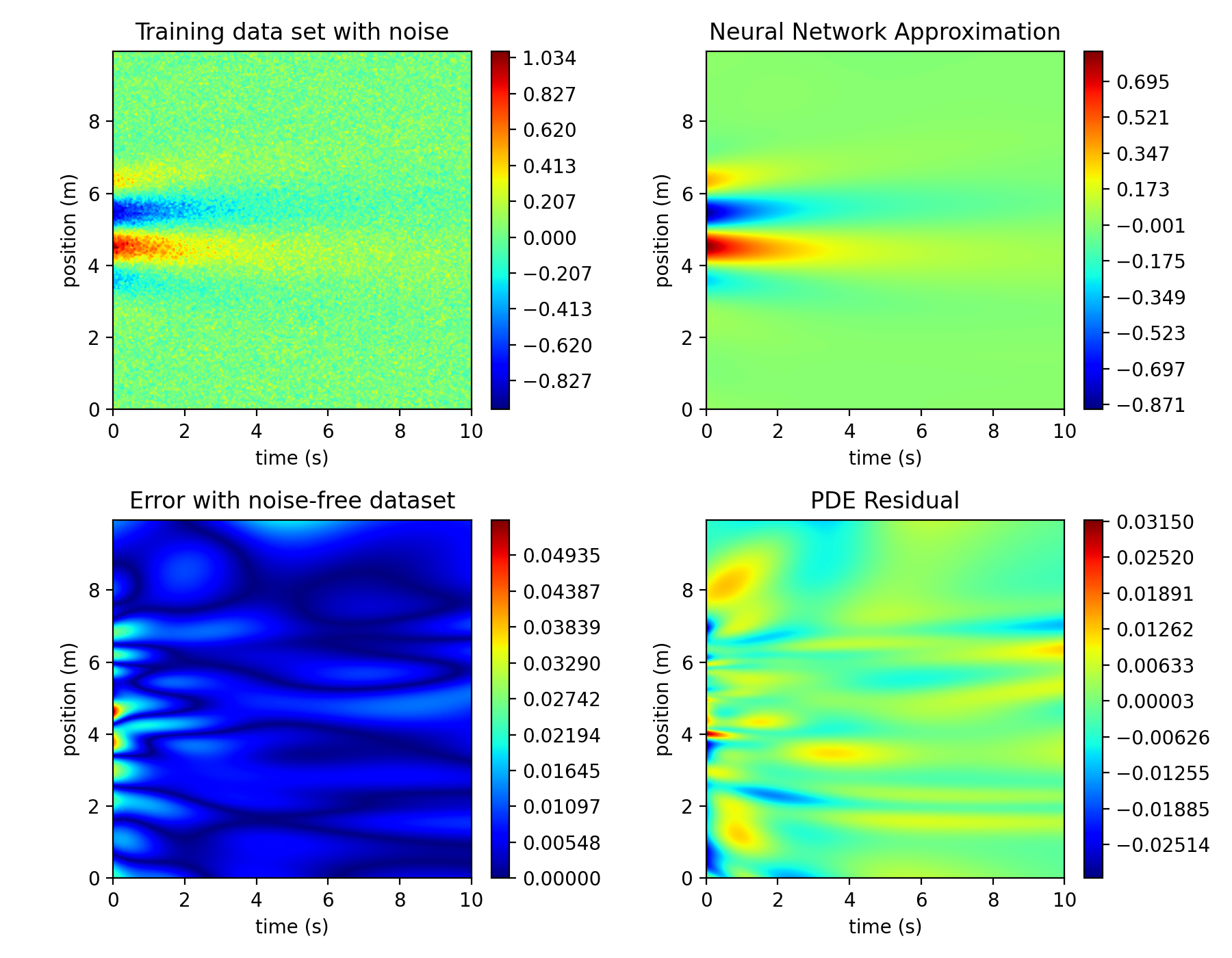}
    \caption{Learned solution for the second heat equation data set with $100$\% noise. 
    The upper left sub-plot shows the noisy data set.
    The upper right sub-plot depicts $U$ after training.
    The bottom left sub-plot shows the absolute error between $U$ and the noise-free data set.
    The bottom right sub-plot depicts the PDE residual on the problem domain, $\Omega$.}
    \label{fig:Heat_Exp_Sine}
\end{figure}

$~$

We train both networks for $2,000$ Epochs using the \texttt{Adam} optimizer, with a learning rate of $0.001$.
After training, \texttt{PDE-READ} uses its sparse regression algorithm to identify candidate PDEs. 
The highest-ranking candidate PDE is

$$D_t U = -(0.040518)(D_{x}^2 U),$$

which matches form of the heat equation (equation \ref{eq:Heat}).
Moreover, the coefficient in the learned PDE has a relative error of approximately $20$\% with respect to that in the true PDE (in this case, $\alpha = 0.05$).
Figure \ref{fig:Heat_Exp_Sine} depicts the learned solution along with residual and error plots.

\begin{figure}[!hbt]
    \centering
    \includegraphics[width=.6\linewidth]{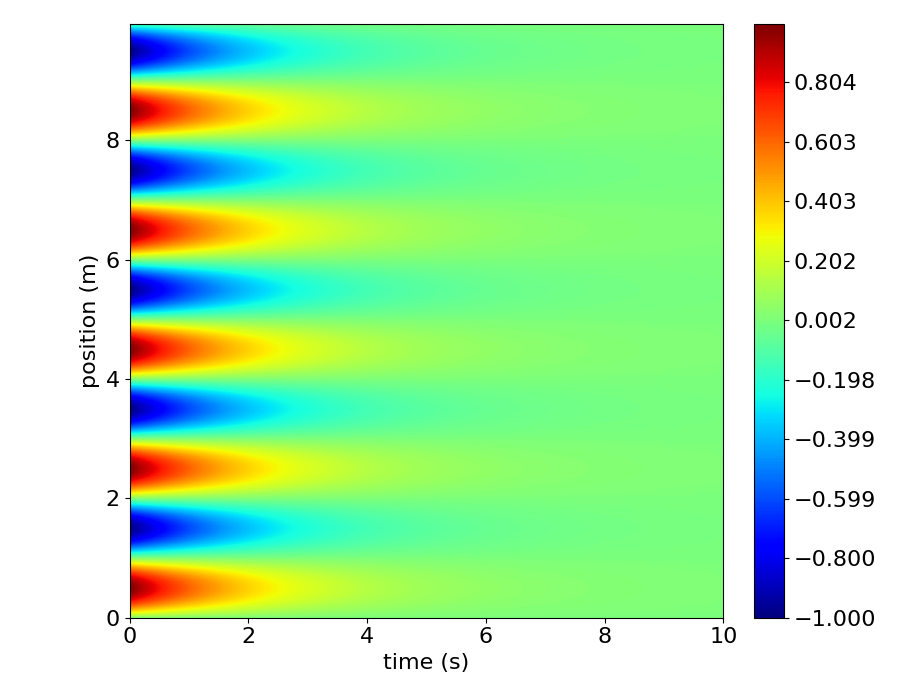}
    \caption{Second Heat equation data set (noise-free).}
    \label{fig:Heat_Sine_Dataset}
\end{figure}

$~$

\noindent {\bf Second data set:} For the second heat equation data set, we use Chebfun to solve the heat equation on $\Omega$ with the following initial condition:

$$U(0, x) = \sin \left( 5 x \left( \frac{2 \pi}{10} \right) \right) \quad\quad x \in (0, 10).$$

Figure \ref{fig:Heat_Sine_Dataset} depicts the noise-free data set.
For this experiment, the {noisy and limited} data set contains $10,000$ data points ($N_{Data} = 10,000$) and $100$\% noise. 
Further, we use $10,000$ collocation points ($N_{Coll} = 10,000$).

$~$

We then train both networks for $2,000$ Epochs using the \texttt{Adam} optimizer, with a learning rate of $0.001$.
After training, \texttt{PDE-READ} uses its sparse regression algorithm to identify candidate PDEs. 
The highest-ranking candidate PDE is

\begin{equation} D_t U = -(0.459) (U). \label{eq:Heat:Wrong_Eq} \end{equation}

Interestingly, this does not match our form of the heat equation (equation \ref{eq:Heat}).
\texttt{PDE-READ} did not fail, however, as it learned a valid PDE for this data set.
In particular, in section \ref{sub_sub_sec:Heat_Discuss}, we demonstrate that the solution for this problem satisfies the identified PDE (with a slightly different coefficient).
This result highlights a fundamental problem with PDE discovery: it is an ill-posed problem. 
In particular, multiple PDEs can describe the same data set.
Given that the identified PDE is lower order than the heat equation, this result implies that \texttt{PDE-READ} learns the simplest (in the sense that lower order is simpler) PDE that a data set satisfies. 
We discuss this point in detail in section \ref{sub_sub_sec:Heat_Discuss}.

\subsection{Burgers' Equation} \label{sub_sec:Burgers}

Burgers' equation arises in many technological contexts, including Fluid Mechanics, Nonlinear Acoustics, Gas Dynamics, and Traffic Flow \cite{basdevant1986spectral}. 
It takes the following form:

\begin{equation} D_{t} U = \nu (D_{x}^2 U) - (U) (D_{x} U). \label{eq:Burgers} \end{equation}

Here, $U(t, x)$ represents the velocity of a fluid at time $t$ and position $x$.
$\nu > 0$ is the {\it diffusion coefficient}.
Burgers' equation is a non-linear second-order PDE. 
Significantly, solutions to Burgers' equation can develop shocks (discontinuities) in finite time. 
We test \texttt{PDE-READ} on two Burgers' equation data sets{, both with $\nu = 0.1$}. 
The first exhibits shock formation, while the latter does not. 

$~$

For both data sets, we use periodic boundary conditions and set $\Omega = (0, T] \times S = (0, 10] \times (-8, 8)$.
We partition $S = (-8, 8)$ into $255$ equally-sized sub-intervals, and $(0, T] = (0, 10]$ into $200$ equally-sized sub-intervals.
This partition engenders a grid with $256$ equally-spaced grid lines along the spatial axis, and $201$ along the temporal axis. 
We use Chebfun's \texttt{spin} class to solve Burgers equation on this grid. 
Thus, both data sets contain velocities at a total of $51,456 = 256*201$ data points. 

$~$

For each experiment with Burgers' equation, our library of candidate terms is 

$$\left\{ \left(U\right)^{p_0}\left(D_x U\right)^{p_1}\left(D_x^2 U\right)^{p_2} : p_0, p_1, p_2 \in \mathbb{N}_{0} \text{ and } p_0 + p_1 + p_2 \leq 2 \right\}.$$

\begin{figure}[!hbt]
    \centering
    \includegraphics[width=.6\linewidth]{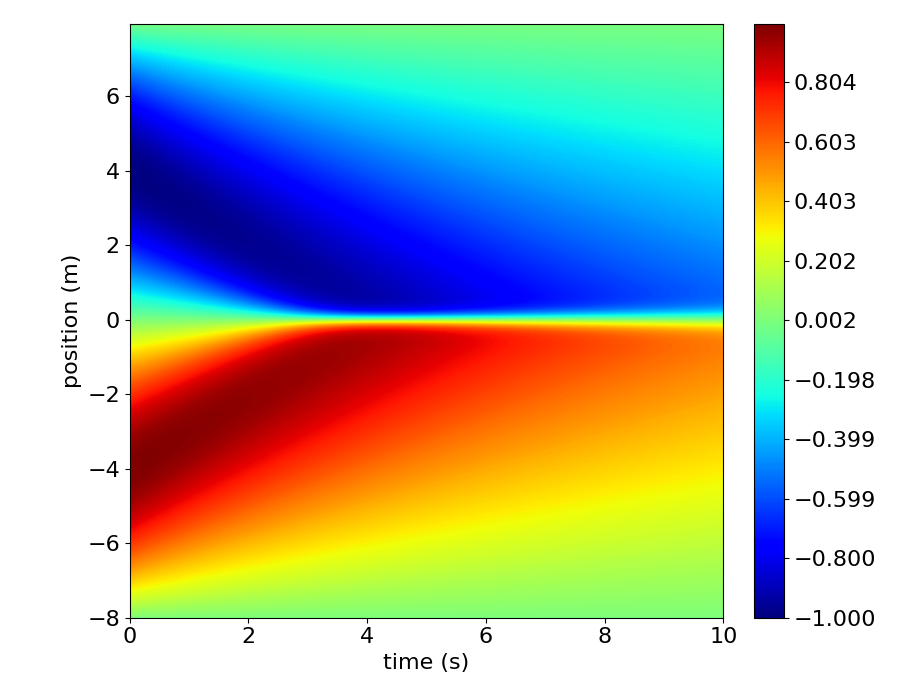}
    \caption{First Burgers' equation data set (noise-free).}
    \label{fig:Burgers_Sine_Dataset}
\end{figure}

\noindent {\bf First data set:} To generate the first data set, we use Chebfun to solve Burgers' equation on $\Omega$ with the following initial conditions:

$$U(0, x) = -\sin\left( \frac{\pi x}{8} \right) \quad\quad x \in (-8, 8).$$

Figure \ref{fig:Burgers_Sine_Dataset} depicts the noise-free data set.
We use this data set to demonstrate \texttt{PDE-READ}'s robustness to sparsity and noise.

$~$

\begin{figure}[!hbt]
    \centering
    \includegraphics[width=\linewidth]{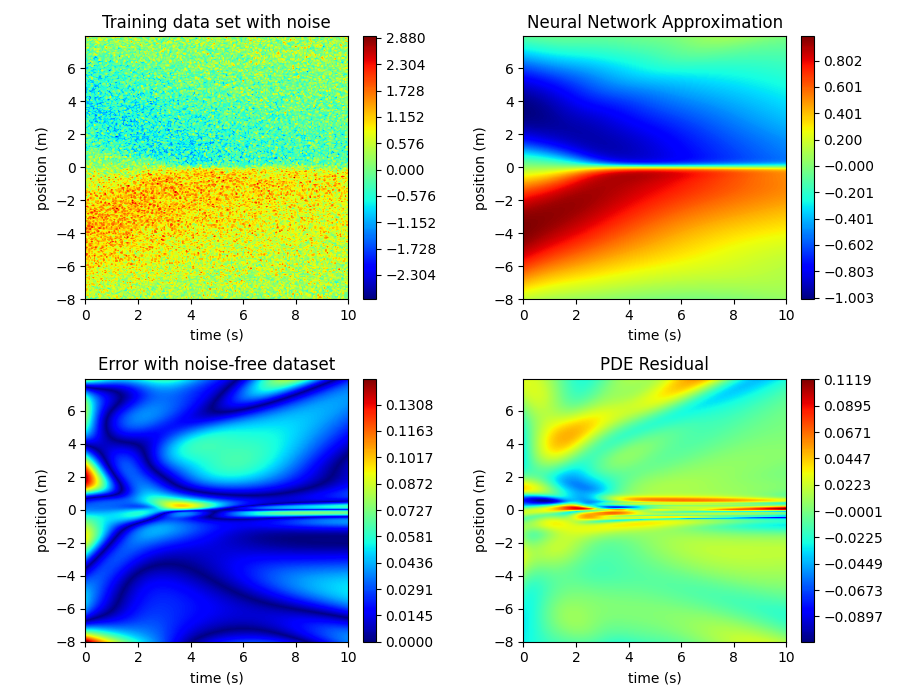}
    \caption{Learned solution for the first Burgers' equation data set with $100$\% noise and velocities at $10,000$ data points. 
    The upper left sub-plot shows the noisy data set. 
    The upper right sub-plot depicts $U$ after training. 
    The bottom left sub-plot shows the absolute error between $U$ and the noise-free data set.
    The bottom right sub-plot depicts the PDE residual on the problem domain, $\Omega$.}
    \label{fig:Burgers_Sine:100}
\end{figure}

First, we train \texttt{PDE-READ} on a {noisy and limited} data set with $10,000$ data points ($N_{Data} = 10,000$) and $100$\% noise. 
We train both networks for $1,500$ epochs using the \texttt{Adam} optimizer, with a learning rate of $0.001$, followed by $100$ epochs of the LBFGS optimizer, with a learning rate of $0.05$.
After training, \texttt{PDE-READ} uses its sparse regression algorithm to identify candidate PDEs. 
The highest-ranking candidate PDE is

$$D_{t} U = 0.086747(D_{x}^2 U) - 0.939752(U)(D_{x} U),$$

which matches our form of Burgers' equation (equation \ref{eq:Burgers}).
The two coefficients exhibit a relative error of about $10$\% with respect to those in the true PDE (equation \ref{eq:Burgers}).
Figure \ref{fig:Burgers_Sine:100} depicts the learned solution along with residual and error plots.

$~$

Next, we repeat the previous experiment with just $4,000$ data points and $100$\% noise. 
We train both networks using $1500$ epochs of \texttt{Adam}, with a learning rate of $0.001$, followed by $100$ epochs of \texttt{LBFGS}, with a learning rate of $0.1$.
After training, the highest-ranking candidate PDE is 

$$D_{t} U = 0.060935(D_{x}^2 U) - 0.624332(U)(D_{x} U),$$

which matches our form of the of Burgers' equation (equation \ref{eq:Burgers}).

$~$

Next, we repeat the previous experiment with $2,000$ data points and $60$\% noise.\footnote{This is the highest noise level at which \texttt{PDE-READ} can reliably identify Burger's equation using just $2,000$ data points}
For this experiment, use $20,000$ collocation points ($N_{Coll} = 20,000$).
We train both networks for $2,000$ epochs using the \texttt{Adam} optimizer, with a learning rate of $0.001$.
After training, the highest-ranking candidate PDE is

$$D_{t} U = 0.051674 (D_{x}^2 U) - 0.666797(U)(D_{x} U).$$

Once again, \texttt{PDE-READ} successfully identifies our form of Burgers' equation (equation \ref{eq:Burgers}).

$~$

Finally, we repeat the previous experiment with just $1000$ data points and $40$\% noise.\footnote{This is the highest noise level at which \texttt{PDE-READ} can reliably identify Burger's equation using just $1,000$ data points}
We again use $20,000$ collocation points ($N_{Coll} = 20,000$).
We train both networks for $2,000$ epochs using the \texttt{Adam} optimizer, with a learning rate of $0.001$.
After training, the highest-ranking candidate PDE is

$$D_{t} U = 0.083974(D_{x}^2 U) - 0.717924(U)(D_{x} U),$$

which matches our form of Burgers' equation (equation \ref{eq:Burgers}).
Notably, however, in each of the three previous experiments, the relative error between the coefficients in the identified PDE and those in the true PDE (equation \ref{eq:Burgers}) is about $40$\%.
We discuss these results in more detail in section \ref{sub_sub_sec:Burgers_Discuss}.
Table \ref{table:Burgers} summarizes the results of our experiments with the first Burgers' equation data set.

\begin{table}
\centering
\begin{tabularx}{.9\linewidth}{c|c|Y}
    Noise Level & Number of Data Points & Identified PDE \\
    \hline
    & & \\[-8pt]
    $100$ & $10,000$ & $D_{t} U = 0.086747(D_{x}^2 U) - 0.939752(U)(D_{x} U)$ \\[2pt]
    $100$ & $4,000$ & $D_{t} U = 0.060935(D_{x}^2 U) - 0.624332(U)(D_{x} U)$ \\[2pt]
    $60$ & $2,000$ & $D_{t} U = 0.051674 (D_{x}^2 U) - 0.666797(U)(D_{x} U)$ \\[2pt]
    $40$ & $1,000$ & $D_{t} U = 0.083974(D_{x}^2 U) - 0.717924(U)(D_{x} U)$
\end{tabularx}
\caption{Results for the first Burgers' equation data set. The true PDE in these experiments is $D_{t} U = 0.1 (D_{x}^2 U) - (U)(D_{x}U)$.}
\label{table:Burgers}
\end{table}

$$~$$

\begin{figure}[!hbt]
    \centering
    \includegraphics[width=.6\linewidth]{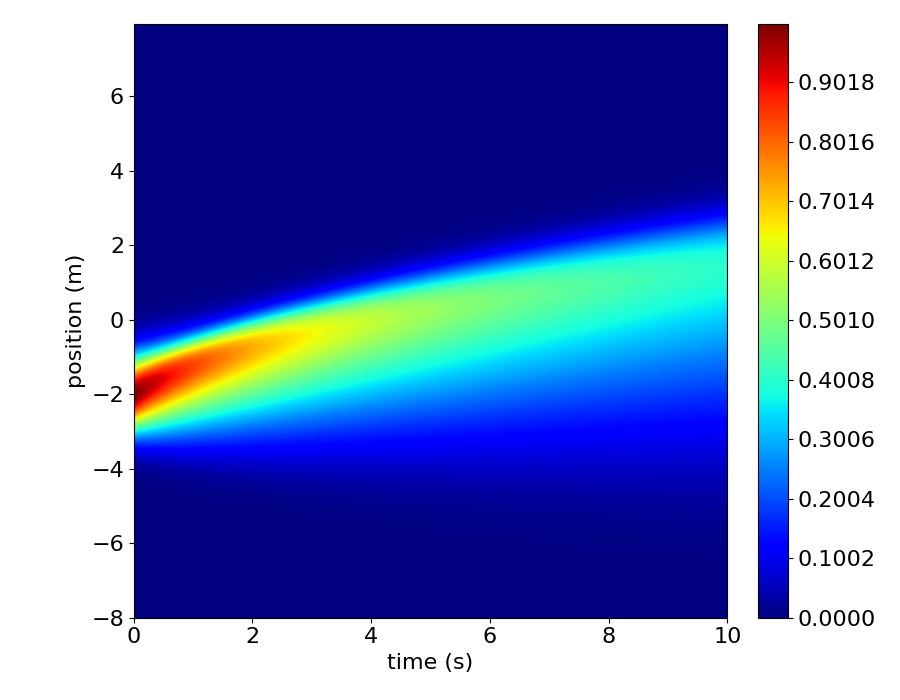}
    \caption{Second Burgers' equation data set without noise.}
    \label{fig:Burgers_Exp_Dataset}
\end{figure}

\noindent {\bf Second data set:} To generate the second data set, we use Chebfun's \texttt{spin} class to solve Burgers' equation on $\Omega$ with periodic boundary conditions and the following initial conditions:

$$U(0, x) = -\exp\left( -(x + 2)^2 \right).$$

Figure \ref{fig:Burgers_Exp_Dataset} depicts the noise-free data set.

\begin{figure}[!hbt]
    \centering
    \includegraphics[width=\linewidth]{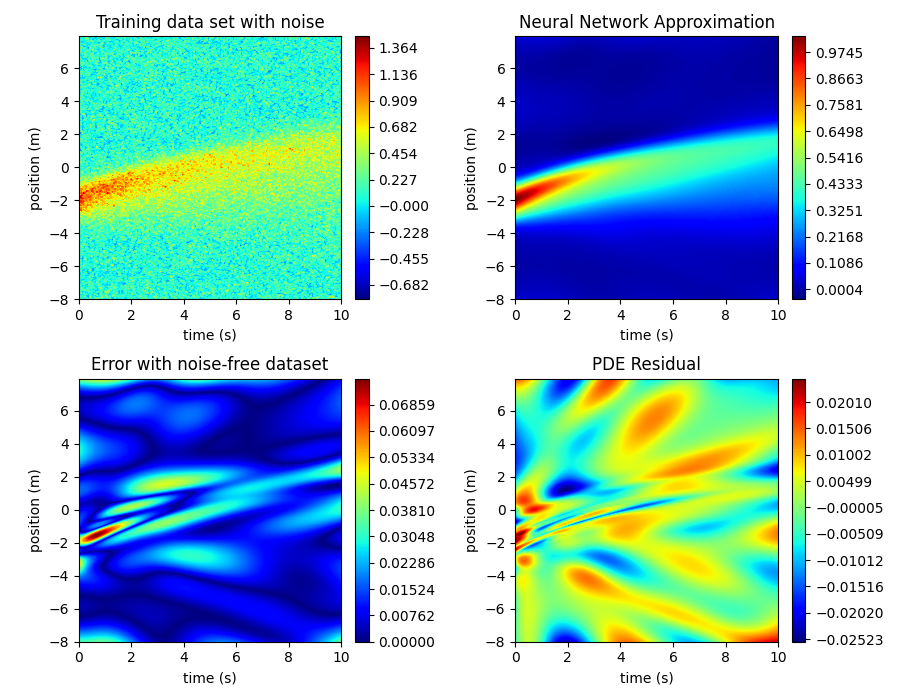}
    \caption{Learned solution for the second Burgers' equation data set with $100$\% noise and velocities at just $4,000$ data points. 
    The upper left sub-plot shows the noisy data set. 
    The upper right sub-plot depicts $U$ after training solution.
    The bottom left sub-plot shows the absolute error between $U$ and the noise-free data set.
    The bottom right sub-plot depicts the PDE residual on $\Omega$.}
    \label{fig:Burgers:4000DP:100}
\end{figure}

$~$

We then create a {noisy and limited} data set with just $4,000$ data points and $100$\% noise.
For this experiment, we use $20,000$ collocation points ($N_{Coll} = 20,000$).
We train both networks for $2,000$ epochs using the \texttt{Adam} optimizer, with a learning rate of $0.001$, followed by $20$ epochs using the \texttt{LBFGS} optimizer, with a learning rate of $0.1$.
After training, \texttt{PDE-READ} uses its sparse regression algorithm to identify candidate PDEs. 
The highest-ranking candidate PDE is

$$D_{t} U = 0.094168(D_{x}^2 U) - 1.065668(U)( D_{x} U),$$

which matches our form of Burgers' equation (equation \ref{eq:Burgers}).
In this case, the coefficients closely match those in the true PDE (equation \ref{eq:Burgers}).
Figure \ref{fig:Burgers:4000DP:100} depicts the learned solution along with residual and error plots.

\subsection{Allen-Cahn Equation}
\label{sub_sec:Allen-Cahn}

\begin{figure}[!hbt]
    \centering
    \includegraphics[width=.6\linewidth]{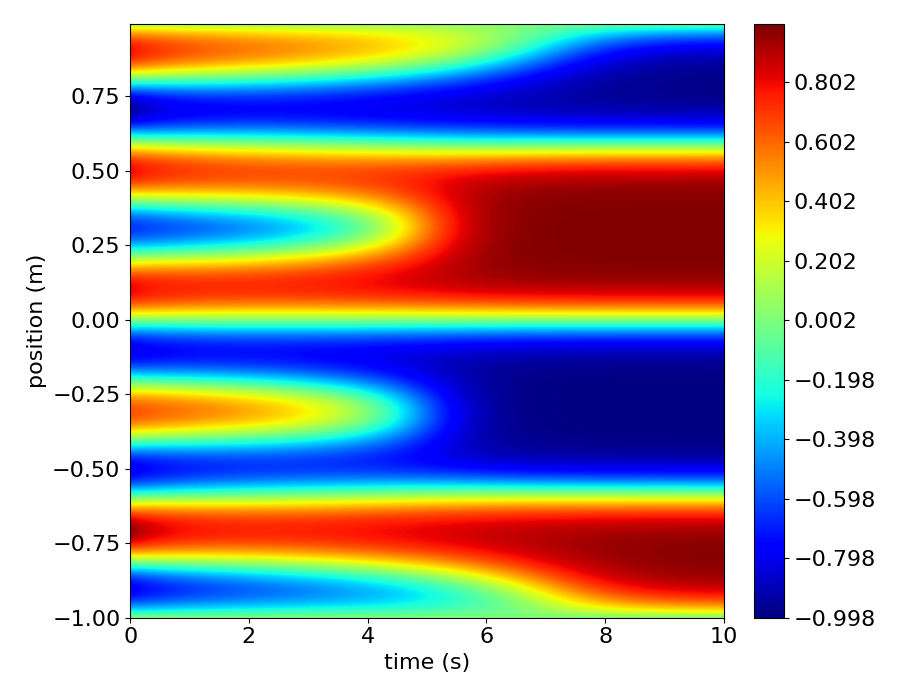}
    \caption{Noise-free Allen-Cahn data set.}
    \label{fig:Allen_Cahn_Dataset}
\end{figure}

The Allen-Cahn equation is a non-linear reaction-diffusion equation which describes phase separation of multi-component metal alloys \cite{allen1979microscopic}.
This second-order equation takes the following form:

\begin{equation} D_{t}U = \varepsilon (D_{x}^2 U) - U^3 + U. \label{eq:Allen-Cahn} \end{equation}

Here, $U$ represents the concentration of one of the constituent metals of the alloy, and $\varepsilon > 0$ is a small constant which characterizes the size of the boundary between the constituent metals.
In our case, $\varepsilon = 0.003$. 

$~$

We test \texttt{PDE-READ} on a the Allen-Cahn equation.
For these experiments, $\Omega = (0, T] \times S = (0, 10] \times (-1, 1)$.
We solve the Allen-Cahn equation on $\Omega$ subject to the initial condition 

$$ U(0, x) = 0.2 \sin \left( 2 \pi x \right)^5 + 0.8 \sin \left(5\pi x \right) \quad\quad x \in (-1, 1). $$

To do this, we partition $S = (-1, 1)$ into $255$ equally-sized sub-intervals, and $(0, T] = (0, 10]$ into $200$ equally-sized sub-intervals. 
This partition engenders a grid with $256$ equally-spaced grid lines along the spatial axis, and $201$ along the temporal axis. 
We use Chebfun's \texttt{spin} class to solve the Allen-Cahn equation on this grid.
The data set contains $51,456 = 256*201$ data points. 
Figure \ref{fig:Allen_Cahn_Dataset} depicts the noise-free data set. 

$~$

We then generate three {noisy and limited} data sets with $50$\%, $75$\%, and $100$\% noise, respectively. 
Each data set has $10,000$ data points. 
In each experiment, we train both networks for $1000$ epochs using the \texttt{Adam} optimizer, with a learning rate of $0.001$, followed by $100$ epochs using the \texttt{LBFGS} optimizer, with a learning rate of $0.1$. 

$~$

After training, \texttt{PDE-READ} uses its sparse regression algorithm to identify candidate PDEs. 
For these experiments, our library of candidate PDE terms is 

$$\left\{ \left(U\right)^{p_0}\left(D_x U\right)^{p_1}\left(D_x^2 U\right)^{p_2} : p_0, p_1, p_2 \in \mathbb{N}_{0} \text{ and } p_0 + p_1 + p_2 \leq 3 \right\}.$$

The highest-ranking candidate PDE in the $50$\% noise experiment is

$$ D_t U = 0.002040(D_x^2 U) + 0.690463(U) - 0.652140(U)^3. $$

The highest-ranking candidate PDE in the $75$\% noise experiment is
$$ D_t U = 0.000958(D_x^2 U) + 0.323689(U) - 0.285130(U)^3. $$

The highest-ranking candidate PDE in the $100$\% noise experiment is 

$$ D_t U = 0.001015(D_x^2 U) + 0.281007(U) - 0.211380(U)^3. $$ 

\begin{figure}[!hbt]
    \centering
    \includegraphics[width=\linewidth]{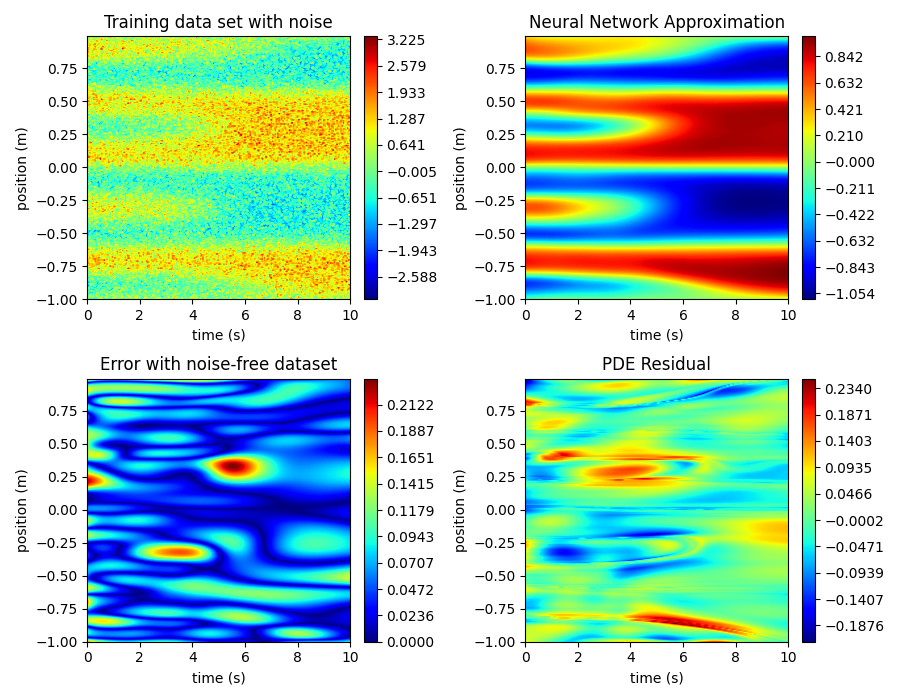}
    \caption{Learned solution for the Allen-Cahn equation data set with $100\%$ noise. 
    The upper left sub-plot shows the noisy data set. 
    The upper right sub-plot depicts $U$ after training.
    The bottom left sub-plot shows the absolute error between $U$ and the noise-free data set.
    The bottom right sub-plot depicts the PDE residual on $\Omega$.}
    \label{fig:Allen_Cahn:100}
\end{figure}

All three identified PDEs match our form of the Allen-Cahn equation (equation \ref{eq:Allen-Cahn}). 
The relative error of the coefficients in the identified PDE with respect to those in the true PDE is roughly $35$\% in the $50$\% noise experiment, $70$\% in the $75$\% noise experiment, and $75$\% in the $100$\% noise experiment.
We discuss these results in section \ref{sub_sub_sec:AC_Discuss}.
Figure \ref{fig:Allen_Cahn:100} depicts the learned solution in the $100$\% noise experiment along with residual and error plots.
Table \ref{table:AllenCahn} summarizes the results of our experiments with the Allen-Cahn equation.

\begin{table}
\centering
\begin{tabularx}{.9\linewidth}{c|c|Y}
    Noise Level & Number of Data Points & Identified PDE \\
    \hline
    & & \\[-8pt]
    $50$ & $10,000$ & $D_t U = 0.002040(D_x^2 U) + 0.690463(U) - 0.652140(U)^3$ \\[2pt]
    $75$ & $10,000$ & $D_t U = 0.000958(D_x^2 U) + 0.323689(U) - 0.285130(U)^3$ \\[2pt]
    $100$ & $10,000$ & $D_t U = 0.001015(D_x^2 U) + 0.281007(U) - 0.211380(U)^3$
\end{tabularx}
\caption{Results for the Allen-Cahn equation. For these experiments, the true PDE is $D_{t} U = 0.003(D_{x}^2 U) - U^3 + U$.}
\label{table:AllenCahn}
\end{table}

\subsection{Korteweg–De Vries Equation}
\label{sub_sec:KdV}

\begin{figure}[!hbt]
    \centering
    \includegraphics[width=.6\linewidth]{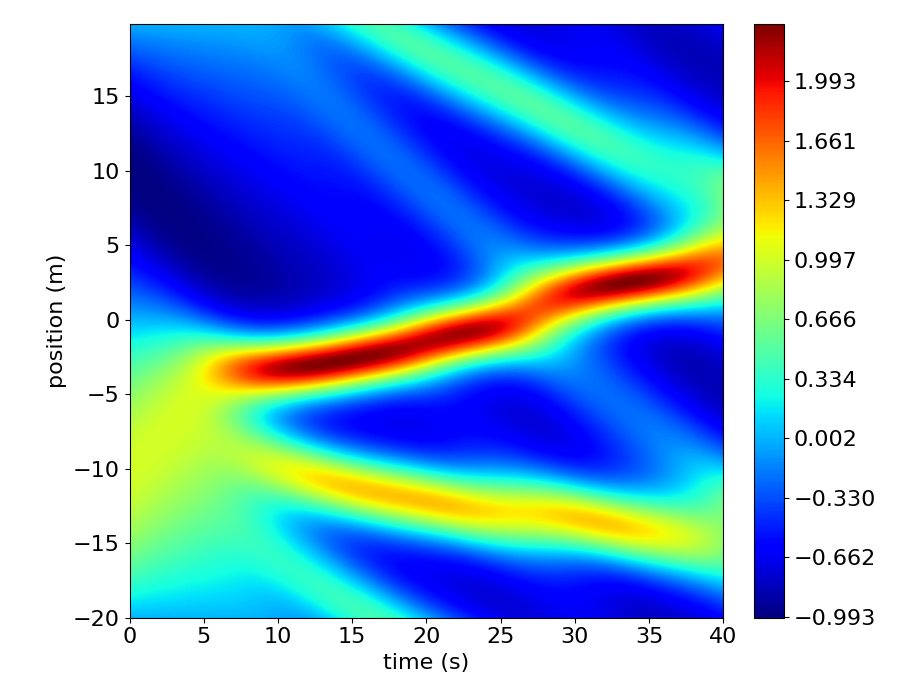}
    \caption{Noise-free KdV equation data set.}
    \label{fig:KdV_Dataset}
\end{figure}

The Korteweg-De Vries (KdV) equation describes the evolution of one-dimensional waves in various settings, including shallow-water conditions.
This third order equation takes the following form:

\begin{equation} D_{t} U = -(U)(D_{x} U) - D_{x}^3 U. \label{eq:KdV} \end{equation}

Here, $U(x, t)$ represents the wave amplitude at time $t$ and position $x$.

$~$

We test \texttt{PDE-READ} on the KdV equation.
For these experiments, $\Omega = (0, T] \times S = (0, 40] \times (-20, 20)$.
We use Chebfun to solve the KdV equation on $\Omega$ subject to periodic boundary conditions and the following initial condition:

$$U(0, x) = -\sin \left( \frac{\pi x}{20} \right) \quad\quad x \in (-20, 20).$$

We partition $S = (-20, 20)$ into $511$ equally-sized sub-intervals, and $(0, T] = (0, 40]$ into $200$ equally-sized sub-intervals.
This partition engenders a grid with $512$ equally-spaced grid lines along the spatial axis, and $201$ along the temporal axis. 
We use Chebfun's \texttt{spin} class to solve the KdV equation on this grid. 
The resulting data set contains $102,912 = 512*201$ data points. 
Figure \ref{fig:KdV_Dataset} depicts the noise-free KdV data set. 

\begin{figure}[!hbt]
    \centering
    \includegraphics[width=\linewidth]{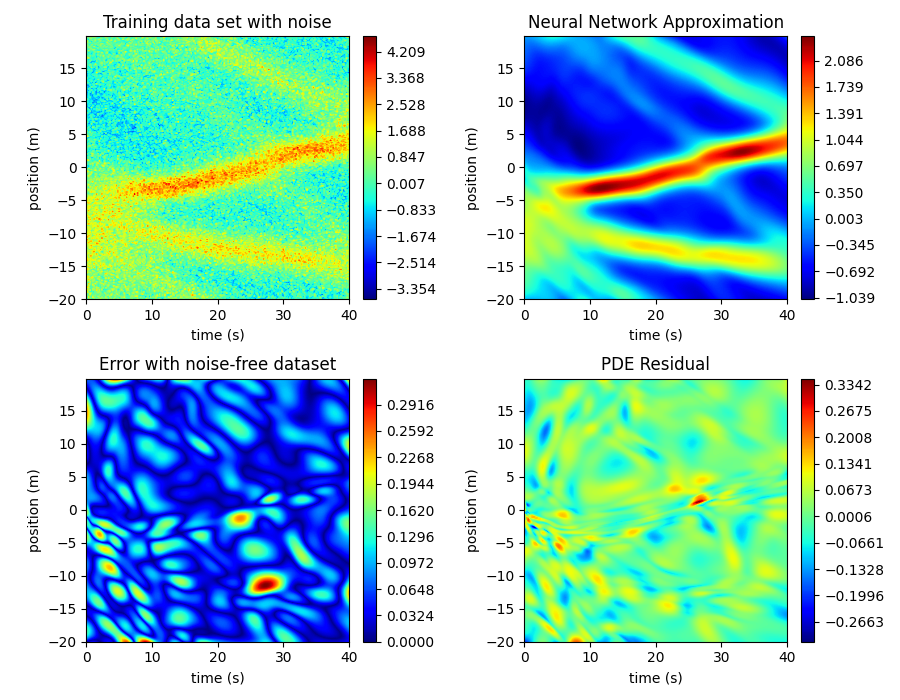}
    \caption{Learned solution for the KdV equation data set with $100\%$ noise. 
    The upper left sub-plot shows the noisy data set. 
    The upper right sub-plot depicts $U$ after training.
    The bottom left sub-plot shows the absolute error between $U$ and the noise-free data set.
    The bottom right sub-plot depicts the PDE residual on $\Omega$.}
    \label{fig:KdV:100}
\end{figure}

$~$

We then generate four {noisy and limited} data sets with $10$\%, $60$\%, $75$\%, and $100$\% noise, respectively.
Each data set has $10,000$ data points. 
In these experiments, we dramatically increase the size of the library to demonstrate the robustness of our sparse regression algorithm with respect to this hyperparameter.

$~$

For the $10$\% and the $60$\% noise experiment, we expand our library of candidate PDE terms to 

$$\left\{ \left(U\right)^{p_0}\left(D_x U\right)^{p_1}\left(D_x^2 U\right)^{p_2}\left(D_x^3 U\right)^{p_3} : p_0, p_1, p_2, p_3 \in \mathbb{N}_{0} \text{ and } p_0 + p_1 + p_2 + p_3 \leq 5 \right\}.$$

For the $10$\% noise experiment, we train both networks for $2,000$ epochs using the \texttt{Adam} optimizer, with a learning rate of $0.001$, followed by $200$ epochs of \texttt{LBFGS}, with a learning rate of $0.1$.
After training, \texttt{PDE-READ} uses its sparse regression algorithm to identify candidate PDEs. 
The highest-ranking candidate PDE is

$$D_{t} U = -0.993144(D_{x}^3 U) - 0.983580(U)( D_{x} U),$$

which matches our form of the KdV equation (equation \ref{eq:KdV}).
The coefficients have a relative error of about $1$\% with respect to the corresponding values in the true PDE, equation \ref{eq:KdV}. 

$~$

For the $60$\% noise experiment, we train both networks for $2,000$ epochs using the \texttt{Adam} optimizer, with a learning rate of $0.001$, followed by $200$ epochs using the \texttt{LBFGS} optimizer, with a learning rate of $0.1$.
After training, the highest-ranking candidate PDE is 

$$ D_{t} U = -0.799669(D_{x}^3 U) - 0.843654(U) (D_{x} U).$$

$~$

For the $75$\% and $100$\% noise experiments, our library of candidate PDE terms is 

$$\left\{ \left(U\right)^{p_0}\left(D_x U\right)^{p_1}\left(D_x^2 U\right)^{p_2}\left(D_x^3 U\right)^{p_3} : p_0, p_1, p_2, p_3 \in \mathbb{N}_{0} \text{ and } p_0 + p_1 + p_2 + p_3 \leq 3 \right\}.$$

For the $75$\% noise experiment, we train both networks using $1500$ epochs using the \texttt{Adam} optimizer, with a learning rate of $0.001$, followed by $200$ epochs using the \texttt{LBFGS} optimizer, with a learning rate of $0.1$. 
After training, the highest-ranking candidate PDE is

$$ D_t U = -0.323292(D_x^3 U) - 0.400611(U)(D_x U). $$ 

For the $100$\% noise experiment, we train both networks using $3000$ epochs of the \texttt{Adam} optimizer, with a learning rate of $0.001$ for the first $1500$ epochs, $0.0003$ for the next $1000$ epochs and $0.0001$ for the final $500$ epochs.\footnote{The \texttt{LBFGS} optimizer diverges in this experiment. For this experiment, its approximate Hessian matrix is ill-conditioned, causing the optimizer to blow up.}
After training, the highest-ranking candidate PDE is

$$ D_t U = -0.307038(D_x^3 U) - 0.399012(U)(D_x U). $$ 

Thus, \texttt{PDE-READ} correctly identifies the KdV equation in all four experiments.
With that said, the relative error between the coefficients in the learned PDEs and those in equation \ref{eq:KdV} are substantial in the $60$\%, $75$\%, and $100$\% experiments.
Figure \ref{fig:KdV:100} depicts the learned solution in the $100$\% noise experiment along with residual and error plots.
We discuss these results in section \ref{sub_sub_sec:KdV_Discuss}.
Table \ref{table:KdV} summarizes the results of our experiments with the KdV equation: 

\begin{table}
\centering
\begin{tabularx}{.9\linewidth}{c|c|Y}
    Noise Level & Number of Data Points & Identified PDE \\
    \hline
    & & \\[-8pt]
    $10$ & $10,000$ & $D_{t} U = -0.(D_{x}^3 U) - 0.983580(U)( D_{x} U)$ \\[2pt]
    $60$ & $10,000$ & $D_{t} U = -0.799669(D_{x}^3 U) - 0.843654(U) (D_{x} U)$ \\[2pt]
    $75$ & $10,000$ & $D_t U = -0.323292(D_x^3 U) - 0.400611(U)(D_x U) $ \\[2pt]
    $100$ & $10,000$ & $D_t U = -0.307038(D_x^3 U) - 0.399012(U)(D_x U)$
\end{tabularx}
\caption{Results for the KdV equation. For these experiments, the true PDE is $D_{t} U = -(U)(D_x U) - D_{x}^3 U$. }
\label{table:KdV}
\end{table}

\subsection{Klein-Gordon Equation}
\label{sub_sec:KG}

\begin{figure}[!hbt]
    \centering
    \includegraphics[width=.6\linewidth]{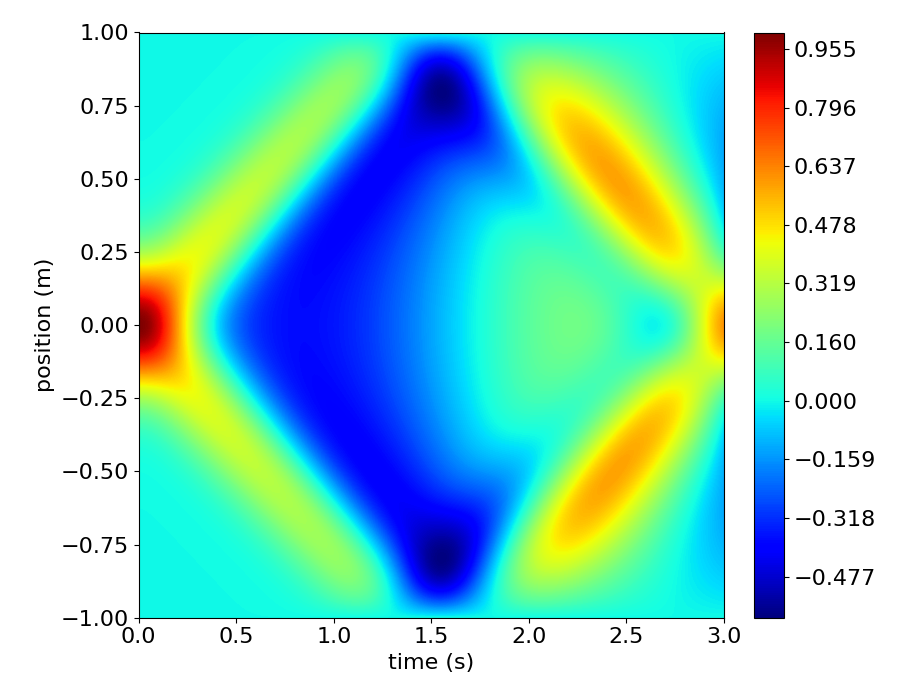}
    \caption{Noise-free Klein-Gordon equation data set.}
    \label{fig:KG_Dataset}
\end{figure}

The Klein-Gordon equation is a relativistic wave equation.
Oskar Klein and Walter Gordon first proposed the equation in 1926 to describe the behavior of electrons in relativistic settings.
In one spatial dimension, this linear second-order equation takes the following form: 

\begin{equation} D_t^2 U = c^2 (D_x^2 U) - \left(\frac{m^2 c^4}{\hbar^2}\right)(U) \label{eq:KG:Unscaled} \end{equation}

where $c$ denotes the speed of light, $\hbar$ denotes Planck's constant, and $m$ is a particle's mass. 
Here, $U$ represents the wave function of a particle. 
Though the KG equation failed to properly describe electrons in relativistic settings, it can model zero-spin particles, such as the Higgs Boson.

$~$

We test \texttt{PDE-READ} on the KG equation.
With appropriate coordinate re-scaling, equation \ref{eq:KG:Unscaled} becomes $D_t^2 U = \gamma D_x^2 U - \mu U$, for some $\gamma, \mu > 0$. 
For these experiments, we set $\gamma = 0.5$ and $\mu = 5$.
Thus, the true PDE is

\begin{equation} D_t^2 U = 0.5 D_x^2 U - 5 U. \label{eq:KG} \end{equation}

Further, we set $\Omega = (0, T] \times S = (0, 3] \times (-1, 1)$.
We use Chebfun's \texttt{Chebop2} class to solve the KG equation on $\Omega$, subject to Dirichlet boundary conditions,

$$U(t, -1) = U(t, 1) = 0 \quad\quad t \in (0, 3],$$

and the initial conditions

$$\begin{aligned} U(0, x) &= \exp\left( -20 x^{2} \right) \\
D_{t} U(0, x) &= 0,\end{aligned}$$

for $x \in (-1, 1)$. 
To construct the noise-free data set, we partition each of $S = (-1, 1)$ and $(0, T] = (0, 10]$ into $200$ equally-sized sub-intervals.
This engenders a regular grid, with $201$ equally-spaced grid lines along both axes. 
We then evaluate Chebfun's solution on the corresponding grid points. 
The resulting data set contains $40,401 = 201*201$ data points. 
Figure \ref{fig:KG_Dataset} depicts the noise-free KG equation data set. 

$~$

We then generate two {noisy and limited} data sets with $50$\% and $100$\% noise, respectively. 
Each data set has $10,000$ data points. 
In each experiment, we train both networks for $2000$ epochs using the \texttt{Adam} optimizer, with a learning rate of $0.001$, followed by $200$, epochs using the \texttt{LBFGS} optimizer, with a learning rate of $0.1$. 
After training, \texttt{PDE-READ} uses its sparse regression algorithm to identify candidate PDEs. 
For these experiments, our library of candidate PDE terms is

$$\left\{ \left(U\right)^{p_0}\left(D_x U\right)^{p_1}\left(D_x^2 U\right)^{p_2} : p_0, p_1, p_2 \in \mathbb{N}_{0} \text{ and } p_0 + p_1 + p_2 \leq 2 \right\}.$$

After training, the highest-ranking candidate PDE in the $50$\% noise experiment is

$$ D_t^2 U = -4.949627(U) + 0.502288(D_x^2 U), $$ 

while the highest-ranking candidate PDE in the $100$\% noise experiment is

$$D_t^2 U = -4.499695(U) + 0.514192(D_x^2 U).$$

\begin{figure}[!hbt]
    \centering
    \includegraphics[width=\linewidth]{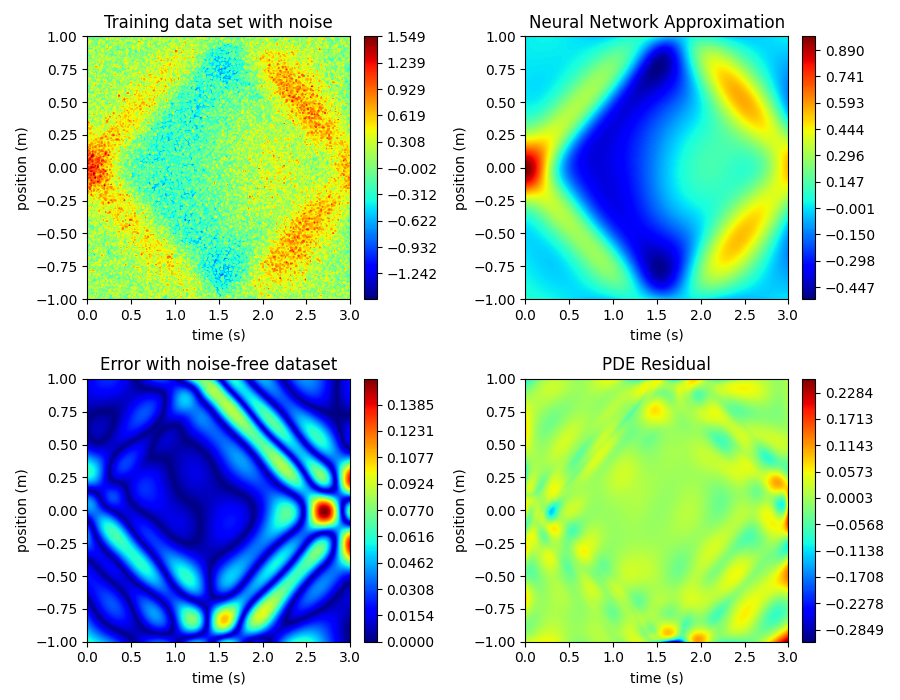}
    \caption{Learned solution for the Klein-Gordon equation data set with $100\%$ noise. 
    The upper left sub-plot shows the noisy data set. 
    The upper right sub-plot depicts $U$ after training.
    The bottom left sub-plot shows the absolute error between $U$ and the noise-free data set.
    The bottom right sub-plot depicts the PDE residual on $\Omega$.}
    \label{fig:KG:100}
\end{figure}

Both of these match our form of the KG equation (equation \ref{eq:KG}). 
Further, the coefficients in the identified PDEs are remarkably close to those in equation \ref{eq:KG}. 
Figure \ref{fig:KG:100} depicts the learned solution in the $100$\% noise experiment along with residual and error plots.
We discuss these results in section \ref{sub_sub_sec:KG_Discuss}.

\subsection{Dynamic Beam Equation}
\label{sub_sec:Beam}

\begin{figure}[!hbt]
    \centering
    \includegraphics[width=.6\linewidth]{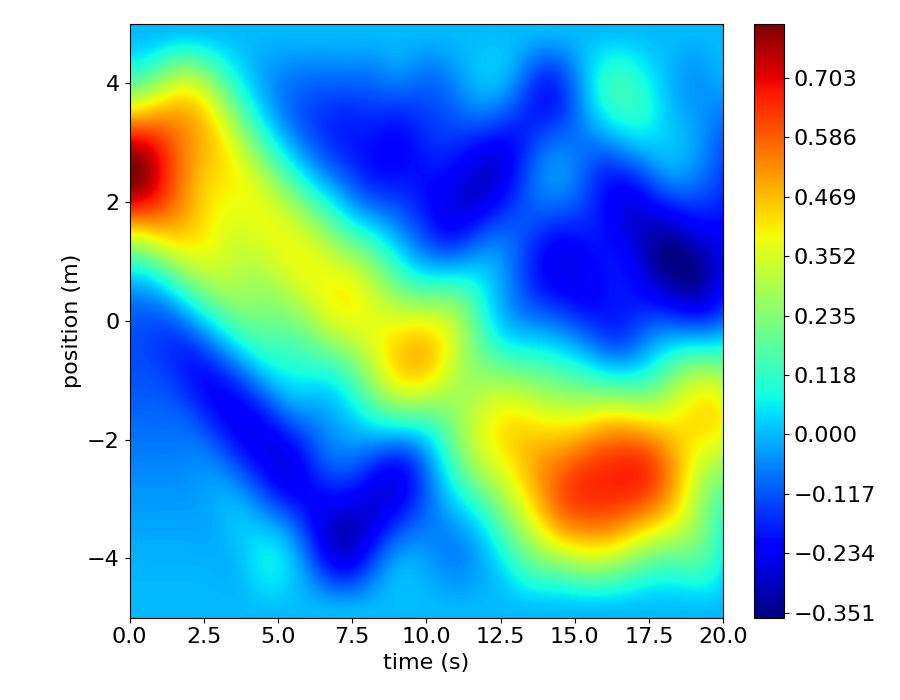}
    \caption{Noise-free dynamic beam equation data set.}
    \label{fig:Beam_Dataset}
\end{figure}

The dynamic beam equation is a linear fourth order PDE that describes the transverse motion of beams with clamped ends. 
It is one of the fundamental equations in Euler-Bernoulli beam theory, a simplification of linear elasticity that is specialized for beams.
For a prismatic, homogeneous beam having no transverse external load, the beam equation is

$$ \mu (D_t^2 U) = -(EI)\left(D_x^4 U\right).$$

Here, $E$, $I$, and $\mu$ represent the modulus of elasticity, second area moment of the beam's cross section, and the beam's mass per unit length, respectively. 

$~$

We test \texttt{PDE-READ} on the dynamic beam equation.
For our experiments, we set $E I = 0.1$ and $\mu = 1$. 
Thus, the true PDE is 

\begin{equation} D_t^2 U = -(0.1)(D_x^4 U). \label{eq:Beam} \end{equation}

Further, we set $\Omega = (0, T] \times S = (0, 20] \times (-5, 5)$.
We use Chebfun's \texttt{Chebop2} class to solve the dynamic beam equation on $\Omega$ subject to fixed end conditions,

$$\begin{aligned} U(t, 5) = U(t, -5) &= 0 \\
D_x U(t, 5) = D_x U(t, -5) &= 0,\end{aligned}$$

and the initial conditions

$$\begin{aligned} U(0, x) &= 1 - \exp\left(-0.5(x - 2.5)^2 \right) \\ 
D_{t} U(0, x) &= 0 \end{aligned}$$

for $x \in (-5, 5)$. 
To construct the noise-free data set, we first partition $S = (-5, 5)$ and $(0, T] = (0, 20]$ into $200$ equally-sized sub-intervals.
This engenders a regular grid with $201$ equally-spaced grid lines along both axes.
We then evaluate Chebfun's solution on the corresponding grid points. 
The resulting data set contains $40,401 = 201*201$ data points. 
Figure \ref{fig:Beam_Dataset} depicts the noise-free dynamic beam equation data set. 

$~$

\begin{figure}[!hbt]
    \centering
    \includegraphics[width=\linewidth]{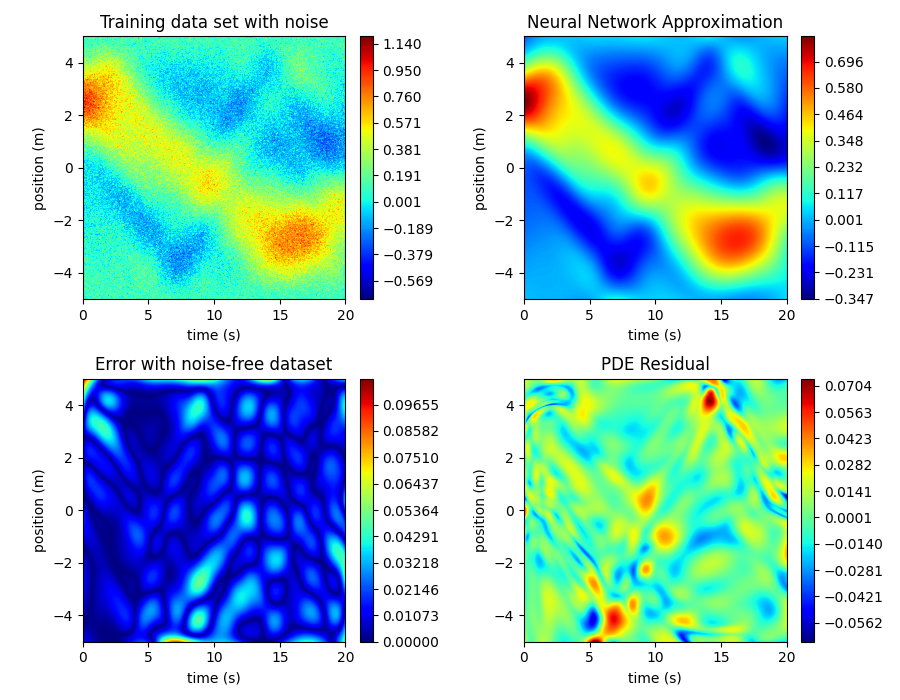}
    \caption{Learned solution for the dynamic beam equation data set with $50\%$ noise. 
    The upper left sub-plot shows the noisy data set. 
    The upper right sub-plot depicts $U$ after training.
    The bottom left sub-plot shows the absolute error between $U$ and the noise-free data set.
    The bottom right sub-plot depicts the PDE residual on $\Omega$.}
    \label{fig:Beam:50}
\end{figure}

We then generate two {noisy and limited} data sets with $25$\% and $50$\% noise, respectively. 
Each data set has $10,000$ data points. 
In each experiment, we train both networks for $2000$ epochs using the \texttt{Adam} optimizer, with a learning rate of $0.001$, followed by $100$ epochs using the \texttt{LBFGS} optimizer, with a learning rate of $0.1$. 
For these experiments, our library of candidate PDE terms is 

$$\left\{ \left(U\right)^{p_0}\left(D_x U\right)^{p_1}\left(D_x^2 U\right)^{p_2}\left(D_x^3 U\right)^{p_3}\left(D_x^4 U\right)^{p_4} : p_0, p_1, p_2, p_3, p_4 \in \mathbb{N}_{0} \text{ and } p_0 + p_1 + p_2 + p_3 + p_4 \leq 2 \right\}.$$

After training, the highest-ranking candidate PDE in the $25$\% noise experiment is

$$D_t^2 U = -0.098640(D_x^4 U),$$ 

while the highest-ranking candidate PDE in the $50$\% noise experiment is

$$D_t^2 U = -0.053314(D_x^4 U).$$

Thus, \texttt{PDE-READ} correctly identifies the dynamic beam equation in both experiments, though the relative error between the lone coefficient is almost $50$\% in the $100$\% noise experiment. 
Figure \ref{fig:Beam:50} depicts the learned solution in the $100$\% noise experiment along with residual and error plots.
We discuss these results in section \ref{sub_sub_sec:Beam_Discuss}.

\section{Discussion} \label{sec:Discussion}

The present section outlines observations made during the forging work. 
The discussion section is split in two sub-sections.
Section \ref{sub_sec:Method_Discuss} discusses the rational behind our methodology, practical considerations for using \texttt{PDE-READ}, theoretical results, and potential generalizations. 
Section \ref{sub_sec:Results_Discuss} discuss the results of our numerical experiments in section \ref{sec:Results}. 

\subsection{Methodology} \label{sub_sec:Method_Discuss}

\subsubsection{Neural Network Approximations to $u$ and $\hat{N}$} \label{sub_sub_sec:NN_Discuss}

\noindent\textbf{The Form of the Hidden PDE:} In this paper, we assume that the system response function, $u$, satisfies a PDE of the form of equation \ref{eq:PDE}.
This assumption is certainly valid for the PDEs in section \ref{sec:Results}, but by no means encapsulates all possible PDEs. 
For example, consider the Kadomtsev-Petviashvili (KP) equation, a two dimensional generalization of the KdV equation:

$$ D_{x} \left( D_{t} U + (U)(D_{x} U) + \varepsilon^2 D_{x}^3 U \right) + \lambda D_{y}^2 U = 0.$$

The KP equation involves two spatial variables, and its only time derivative is mixed with a spatial partial derivative.
Currently, \texttt{PDE-READ} can only learn PDEs of the form of equation \ref{eq:PDE}. 
Thus, \texttt{PDE-READ} can not learn the KP equation.
{Likewise, it can not currently learn a rational PDE (one where $\hat{N}$ is a rational function of $u$ and its derivatives).}
With that said, \texttt{PDE-READ} can be generalized to learn other classes of PDEs.
Determining the efficacy of \texttt{PDE-READ} in these other cases represents a potential area of future research.

$~$

There is a limit to how much we can weaken the form of the hidden PDE. 
For example, a natural generalization of equation \ref{eq:PDE} is 

\begin{equation} \hat{N}(D_{t}^{n} u, u, D_{x} u, \ldots, D_{x}^{M} u) = 0, \label{eq:PDE:Generalize} \end{equation}

for some non-linear map $\hat{N}$. 
Unfortunately, learning such a map is difficult in this form; any neural network approximating $\hat{N}$ can learn the zero map, which perfectly satisfies equation \ref{eq:PDE:Generalize} irrespective of $U$ (meaning that $N$ will not learn $\hat{N}$).
The less general form encapsulated by equation \ref{eq:PDE} eliminates this deficiency, as the neural network approximation $N$ must learn to match the time derivative of $U$. 
This prevents $N$ from learning the zero map, unless $U$ actually is constant in time.

$~$

\noindent\textbf{Rational Activation Functions:} In \texttt{PDE-READ}, $N$ and $U$ are RatNNs. 
However, our implementation of \texttt{PDE-READ} allows the user to change which activation functions $U$ and $N$ use.
In particular, the user can choose to use rational, $\tanh$, or $\sin$ activation functions.\footnote{If $U$ or $N$ use the $\sin$ activation function, then we initialize their weight and biases according to the paper \cite{sitzmann2020implicit}}
In our experiments, however, RatNNs seem to outperform their $\tanh$ and $\sin$ counterparts. 
For example, with $4,000$ data points, \texttt{PDE-READ}, using $\tanh$ activation function, can only reliably identify Burgers' equation if the data set contains $50$\% noise, or less; just half of what \texttt{PDE-READ} using RatNNs can handle. 
This result makes sense, given the findings in \cite{boulle2020rational}. 
In particular, a RatNN is effectively an extremely high order rational function, theoretically allowing the network to take advantage of the powerful approximation properties of functions.
For these reasons, \texttt{PDE-READ} uses RatNNs by default.

$~$

We took great care to implement our rational activation functions as efficiently as possible. 
In particular, we use Homer's rule to minimize the number of computations needed to evaluate the numerator and denominator polynomials.
We also experimented with dozens of small tweaks to improve the speed with which the python interpreter evaluates them.
Despite this, our RatNNs run slow.
In particular, a forward pass through one of our RatNNs takes about an order of magnitude longer than one through a $\tanh$ or $\sin$ network having the same architecture. 
However, we fully believe our RatNNs run slowly because they are written in Python, and not because RatNNs are fundamentally slow.

$~$

A type $(3, 2)$ rational function is fundamentally simpler than $\tanh$ or $\sin$, in that it requires far fewer arithmetic operations to evaluate.
The key difference is that PyTorch's $\tanh$ and $\sin$ functions are ``built-in" and fully compiled, while our rational activation function has to be interpreted.
When Python encounters $\tanh$ or $\sin$, it evaluates those functions using a compiled routine.
This allows Python to evaluate the functions very efficiently, notwithstanding their fairly complicated nature.
By contrast, when Python encounters our rational activation function, it has to interpret our python code line-by-line before evaluating the function.
Most of the run-time goes into interpreting the lines of code, rather than executing computations.
As a result, our rational activation functions evaluate slowly, despite their relatively simple nature. 

$~$

The problem gets even worse when we differentiate RatNNs. 
PyTorch has to use its differentiation rules to evaluate the derivative of our rational activation functions. 
This process takes time. 
By contrast, PyTorch can use built-in routines to rapidly evaluate the derivative of $\tanh$ or $\sin$.
Thus, the run-time penalty associated with using RatNNs increases with the order of the hidden PDE.

$~$

All of this highlights one of the primary limitations of using an interpreted language, like Python, for computationally demanding tasks.
Notwithstanding, these observations reveal a natural solution: Implement our rational activation functions in a fully compiled language, like C++.
Indeed, PyTorch has a C++ back-end for its python front-end. 
Nonetheless, most of our experiments still run in a reasonable amount of time\footnote{Generally a few minutes, and rarely more than an hour or two.} on an E2 high-memory GCP node.
With that said, the computational cost of evaluating our rational activation functions in Python may be unacceptably large when trying to uncover high-order PDEs from large data sets.
As such, we believe our rational activation functions should be implemented in C++ before attempting to use \texttt{PDE-READ} on highly intricate data sets. 

$~$

\noindent\textbf{Collocation Points and Re-selection:} We re-select the collocation points before each epoch because it enhances their regularizing power and increases the computational efficiency of our approach. 
We discuss the latter point in section \ref{sub_sec:Results_Discuss}. 
Here, we focus on the former point: That collocation points regularize $U$, and re-selecting them enhances this effect.
The collocation loss prevents $U$ from simply memorizing the noisy measurements of the system response function. 
$U$ must also learn to satisfy the learned PDE, $N$. 
In other words, the two-network approach introduces an inductive bias encapsulates the understanding that $U$ is a PDE solution into the loss function. 
We fully believe this inductive bias is the biggest strength of the two-network approach, and explains its robustness to sparsity and noise.
Without it, $U$ would be free to over-fit the noisy measurements, leading to poor generalization. 
The collocation points and $N$ help $U$ ``denoise" the measurements and learn the underlying function.

$~$

This naturally raises the question: How many collocation points should we use? 
As many as possible, we believe.
Increasing $N_{Coll}$ makes \texttt{PDE-READ} more robust to noise and sparsity.
If there are only a few collocation points, then $N$ and $U$ can ``conspire" such that $U$ satisfies $N$ just at those points. 
In this case, $U$ is free to over-fit the noisy measurements, and $N$ fails to make $U$ learn a solution to the hidden PDE.
By contrast, if $N_{Coll}$ is sufficiently large, then the easiest way to make the collocation loss small is for $N$ to learn the hidden PDE, and for $U$ to learn a solution to it.
What ``sufficiently large" means is problem dependent, but we believe that having more collocation points than network parameters is ``sufficiently large". 
In our case, that meant using about $10,000$ collocation points (which is what we use in most experiments).
However, we believe \texttt{PDE-READ} should use as many collocation points as is practical.
We discuss further in section \ref{sub_sec:Results_Discuss}.

$~$ 

Re-selecting the collocation points each epoch effectively adds new collocation points without increasing computational demand.
An important property of collocation points is that they are unsupervised in nature; we do not need training data to make them.
Creating new collocation points is as simple as generating new coordinates.
This means that re-selecting collocation points is very cheap.
However, periodic re-selection forces $U$ to satisfy the learned PDE at a different set of points every few epochs.
In essence, re-selection simulates having more collocation point. 
This is the principal reason why we re-select the collocation points.

$~$

\noindent\textbf{Other Regularization Techniques:} Since many of our experiments use relatively small data sets, over-fitting is a major concern.
Re-selecting the collocation points before each epoch dramatically reduces over-fitting, and is one of the defining features of \texttt{PDE-READ}.
However, over-fitting training data is not a novel problem in the deep learning world.
Many schemes already exist to limit over-fitting.
One popular approach is dropout \cite{srivastava2014dropout}, which randomly removes hidden units from the network during each forward pass.
A user-defined constant controls the proportion of hidden units that dropout removes. 
Dropout reduces the network's size during training, and effectively forces the optimizer to optimize an entire ensemble of related networks (namely those networks we can obtain by removing a subset of the original network's neurons).
Both of these effects limit over-fitting.
Further, proponents of dropout argue that it makes the network more robust: The network can not rely on any one hidden unit (since dropout may remove that hidden unit on some forward passes). 
The network learns to give predictive results even when some hidden units are missing.

$~$

In practice, we find that dropout techniques dramatically {\it reduces} the performance of \texttt{PDE-READ}. 
With it, our networks fail to learn the hidden PDE. 
We suspect the networks can not learn an approximation of the solution and its derivatives (with respect to the network inputs) that maintain their predictive power when dropout randomly removes hidden units. 

$~$

We also tried using another popular approach, batch normalization.
This approach applies an affine transformation before each hidden layer, which adjusts the components of a mini-batch of inputs to have mean particular mean and variance (which is learned along with the network's weights and biases).
Proponents of batch normalization argue that it allows the network to learn an optimal scale for each layer in the network; thus helping speed up training and reduce activation function saturation.
In practice, we find that using batch normalization reduces \texttt{PDE-READ}'s performance.
Because of these results, neither dropout nor batch normalization are part of our approach. 

$~$

\subsubsection{Sparse Regression via Recursive Feature Elimination} \label{sub_sub_sec:Sparse_Discuss}

\noindent\textbf{Extraction points:} Automatic differentiation allows us to differentiate $U$ and thus, evaluate the library of PDE candidate terms anywhere in the problem domain, $\Omega$.
This also means we can have as many extraction points as we desire. 
We generate extraction points by sampling a uniform distribution over $\Omega$. 
Our approach keeps the collocation, data, and extraction points independent. 

$~$

Increasing the number of extraction points gives us a better picture of how $N$ behaves throughout $\Omega$. 
Adding extraction points should, therefore, improve how well the solution to equation \ref{eq:LeastSquares} approximates the solution to equation \ref{eq:PolyPDE:Short:2}. 
Thus, $N_{Extract}$ should be as large as is practical. 
Adding extraction points adds rows to the library matrix. 
Since our sparse regression algorithm solves a sequence of progressively sparser least-squares problems, increasing the number of extraction points increases run-time. 
However, in our experience, our sparse regression algorithm runs quickly (a few seconds), even when $N_{Extract}$ is quite large ($20,000$ or more). 
As such, we use $20,000$ extraction points in all our experiments. 

$~$

\noindent\textbf{Alternative Sparse Regression Algorithms:}
Our sparse regression algorithm is one of the greatest strengths of \texttt{PDE-READ}.
In particular, \texttt{PDE-READ} does not work nearly as well if we replace our sparse regression algorithm with STRidge, LASSO, or thresholded LASSO.\footnote{First, use LASSO to find a sparse solution. Determine the features that are zero in the LASSO solution. Discard these features and then find the least-squares solution using the remaining features}
We believe this is because the other algorithms suffer from two main problems: parameter fine-tuning, and undesirable bias. 
With this in mind, let us consider STRidge.
To make STRidge work, we must select a value for $\alpha$. However, if $\alpha$ is too big, then the solution contains unnecessary terms. 
By contrast, if $\alpha$ is too small, then the solution excludes too many terms and has little predictive power. 
\texttt{PDE-READ} with STRidge only identifies the correct PDE if $\alpha$ is in a narrow range. 
Worse yet, the set of appropriate $\alpha$ values appears to be problem dependent. 
If we fine-tune $\alpha$ to identify the correct PDE in one experiment, \texttt{PDE-READ} with STRidge and the same $\alpha$ value will often not identify the correct PDE in another experiment. 
As a result, picking an appropriate threshold is difficult without knowing the correct PDE {\it a priori}. 
We call this the {\it fine-tuning problem}. 
Second, over-fitted least-squares solutions often exhibit large coefficients, which makes them less likely to be thresholded by STRidge. 
In this sense, STRidge rewards over-fitting. 
While appropriate preconditioning can, to some degree, assuage the second issue, the fine-tuning problem is inescapable. 
Both LASSO and thresholded LASSO suffer from similar deficiencies. 

$~$

Avoiding the fine-tuning problem is the principal reason we developed a parameter-free sparse regression algorithm. 
We want a practical algorithm that can work in real-world settings {and} believe a parameter-free sparse regression algorithm brings us {closer to realizing this goal.
With that said, RFE is not the only parameter-free sparse regression algorithm.
Other prominent examples include \emph{Orthogonal Matching Pursuit} (OMP), \emph{Forward Regression Orthogonal Least-Squares} (FROLS), and \emph{Sparse Step-wise Regression} (SSR).

$~$

OMP, first proposed in} \cite{pati1993orthogonal}, {is a general algorithm for constructing the orthogonal projection of a vector, $f$, in a Hilbert space, $\mathcal{H}$, onto the span of a set of vectors, $D \subseteq \mathcal{H}$.
It does so by inductively, and iteratively identifying a sequence of vectors in $D$.
At the $k$th iteration, the algorithm determines the residual between $f$ and the orthogonal projection of $f$, onto the $k$ vectors it has already identified - $x_1, \ldots, x_k \in D$ - and then identifies a vector $x_{k + 1} \in D - \{ x_1, \ldots, x_n \}$, such that the component of the residual in the direction of $x_{k + 1}$ is sufficiently large. 
When restricted to finite-dimensional vector spaces, OMP embodies a powerful, parameter-free technique for identifying a sequence of candidate solutions to $Ax = b$.}

$~$

{FROLS, described in section 3.2 of} \cite{billings2013nonlinear} {, takes a slightly different approach.
The method is primarily concerned with identifying the columns of $A$, in a sparse solution to $Ax = b$, rather than the corresponding components of $x$,
FROLS first finds the column of $A$ with the largest projection in the direction of $b$. 
Next, having identified $s$ such columns of $A$, it finds an orthogonal basis for those columns, and then orthogonalizes the remaining columns of $A$ against that basis.
The algorithm then determines which modified column has the largest projection in the direction of $b$, which becomes the next column FROLS considers.
After identifying $s$ columns of $A$ - $a_{j_1}, \ldots, a_{j_s}$ - the user can solve for the corresponding coefficients by solving the constrained least-squares problem $\text{argmim} \{ \| Ax - b \|_2 : x_i = 0 \text{ if } i \notin \{ j_1, \ldots, j_s \} \}$.}

$~$

{SSR is quite similar to RFE, using the same approach to identify a sequence of candidates, though with a slightly different method for selecting the optimal candidate.
SSR has successfully been employed in system identification applications.
For example, } \cite{boninsegna2018sparse} {utilized it, along with a modification of the \texttt{SINDY} algorithm} \cite{brunton2016discovering} {to identify stochastic differential equations from data.}

$~$

{All four algorithms - RFE, OMP, FROLS, and SSR - are parameter-free sparse regression techniques that have been used extensively. 
We found that RFE works well for our algorithm. 
Testing \texttt{PDE-READ} using other parameter-free techniques, such as OMP, FROLS, or SSR, is a possible area of future research.} 

$~$

Fine-tuning aside, {\texttt{PDE-READ} with our sparse regression algorithm appears to be more robust than \texttt{PDE-READ} with} either STRidge or LASSO.
Consider the first Burgers' equation data set from section \ref{sub_sec:Burgers}. 
\texttt{PDE-READ} with STRidge can not reliably identify Burgers' equation from this data set (and $10,000$ data points) if it contains more than $10$\% noise. 
\texttt{PDE-READ} with LASSO is even worse; it can not identify Burgers' equation if the data set contains more than $7$\% noise. 
This is in stark contrast to \texttt{PDE-READ} with our sparse regression algorithm, which identifies Burgers' equation with up to $100$\% noise using just $4,000$ data points. 

$~$

{Finally, recent work by} \cite{fasel2022ensemble} {demonstrates that ensemble learning, via bagging of the library terms, can enhance existing PDE discovery techniques, such as \texttt{PDE-FIND}.
Exploring the efficacy of ensemble techniques with our approach represents a possible area of future research.}

$~$

\noindent\textbf{Theoretical Examination of Our Sparse Regression Algorithm:}
Let us now explore why \texttt{PDE-READ} is so robust to noise. 
Let $A$ be an $n$ by $m$ matrix and $b \in \mathbb{R}^n$ be a vector. 
We want to find the sparse vector, $x \in \mathbb{R}^m$, which approximately solves $Ax = b$. 
In this section, we consider how our sparse regression algorithm handles this problem. 
Given a candidate solution, $x$, we seek the component, $x_k$, of $x$, which does the least to minimize the least-square residual. 
In other words, we seek

\begin{equation} \text{argmin}_{k} \{ R_{LS}(x - x_k e_k) - R_{LS}(x) \}. \label{eq:argmin:fixed} \end{equation}

As in \cite{guyon2002gene}, we define the least important feature (LIF) of $x$ as the component of $x$ with the smallest magnitude. 
In general, this is not a principled approach, as the LIF may not correspond to the solution of equation \ref{eq:argmin:fixed}. 
The situation changes, however, if we appropriately normalize $A$. 
To begin, $R_{LS}(x)$ is a quadratic function of each component of $x$. 
In particular, 

\begin{equation} R_{LS}(x) = \sum_{i = 1}^{n} \left( \left( \sum_{j = 1}^{m} A_{i,j} x_j \right) - b_i \right)^2. \label{eq:Residual:Component} \end{equation}

Therefore, $R$ is a smooth function of $x_k$. 
Thus, we can use a second-order Taylor approximation of $R$ as a function of $x_k$ to estimate the value of the least-squares residual of $x$ with the $k$th feature removed ($x - x_k e_k$). 
In particular, 

$$R_{LS}(x - x_k e_k) = R_{LS}(x) + \left(\frac{dR_{LS}(x)}{dx_k} \right)x_k + \left(\frac{1}{2}\right)\left(\frac{d^2R_{LS}(x)}{dx_k^2}\right)x_k^2.$$

Since the least-squares solution corresponds to the global minimum of $R$, we must have $dR_{LS}(x)/dx_k = 0$. x
Thus,

$$R_{LS}(x - x_k e_k) - R_{LS}(x) = \left(\frac{1}{2}\right)\left(\frac{d^2R_{LS}(x)}{dx_k^2}\right)x_k^2.$$

Differentiating equation \ref{eq:Residual:Component} gives

$$\frac{d^2R_{LS}(x)}{(dx_k)^2} = \sum_{i = 1}^n A_{i,k}^2 = 2\|A_k\|_2^2.$$

In other words, $(d^2R_{LS}(x)/dx_k^2)$ is twice the square of the $L^2$ norm of $A_k$ (the $k$th column of $A$). 
If we divide the $A_k$ by $\| A_k \|_2$, then each column of the resulting matrix has unit $L^2$ norm. 
If we replace $A$ by this new, normalized matrix, then

$$R_{LS}(x - x_k e_k) - R_{LS}(x) = x_k^2.$$

Thus, in this case, the magnitude of a feature indicates how much removing that feature increases the least-square residual. 
This is exactly what our sparse regression algorithm does. 
The above argument proves the following critical result: 

$~$

{\it {\bf Theorem:} if $x$ is a least-squares solution to $Ax \approx b$, and if each column of $A$ has a unit $L^2$ norm, then the LIF of $x$ solves equation \ref{eq:argmin:fixed}.}

$~$

Notably, this result only holds if we use equation \ref{eq:argmin:fixed} to determine the LIF.
Another approach is to find define $x_k$'s importance as the least-squares residual of the least-squares solution to $Ax \approx b$ subject to $x_k = 0$. 
That is, 

\begin{equation} \text{argmin}_{x} \big\{ \| Ax - b \|_{2} : x_k = 0 \big\}. \label{eq:optimal_import}\end{equation}

For brevity, we call this value the \emph{optimal importance} of $x_k$.
Using optimal importance, the LIF is given by

\begin{equation} \text{argmin}_{k} \Big\{ \text{min}_{x} \big\{ \| Ax - b \|_{2} : x_k = 0 \big\} \Big\}. \label{eq:argmin:optimal} \end{equation}

In equation \ref{eq:argmin:fixed}, we remove the $k$th component but keep the others fixed. 
By contrast, in equation \ref{eq:argmin:optimal}, we remove the $k$th feature but allow the others to change. 
Thus the solution to equation \ref{eq:optimal_import} will not, in general, equal $x - x_k e_k$.
Thus, the LIF under optimal importance may differ from the LIF under our importance metric.
With that said, evaluating a feature's optimal importance requires a least-squares solve. 
Thus, if our library contains $n$ terms, then RFE using optimal importance requires $\Theta(n^2)$ least-squares solves. 
By contrast, RFE using our importance metric requires just $n$ least-squares solves. 
For anything other than a small library, RFE with optimal importance requires significantly more computations than our approach.\footnote{Notably, we can evaluate each feature's optimal importance in parallel.
Thus, RFE with optimal importance could utilize parallel computing.
For a moderately sized library, this would reduce the run-time difference between our approach and one using optimal importance.
Notwithstanding, such an approach would still require dramatically more computations than our approach.
Further, a parallel approach can only go so far and would likely face run-time issues if the library is sufficiently large.}

$~$

As a final point, we can understand our importance metric as an efficient upper bound on optimal importance.
To see this, notice that for a given $k$, 

$$\min_{x} \{ \| Ax - b \|_{2} : x_k = 0 \} \leq \|A(x - x_k e_k) - b \|_{2} = R_{LS}(x - x_k e_k).$$

In other words, a feature's optimal importance is no more than $R_{LS}(x)$, plus its importance under our metric.
Our importance metric therefore gives an upper bound on the a feature's optimal importance.
As such, the LIF under our importance metric is the feature with the smallest upper bound.
Our approach yields a computationally inexpensive approximation to RFE with optimal importance.
Further, as our experiments demonstrate, RFE with our importance metric readily identifies the correct PDE across a diverse set of examples.

\subsection{Results} \label{sub_sec:Results_Discuss}

\noindent\textbf{Picking the Number of Data Points:} In most experiments, we somewhat arbitrarily use $10,000$ data points. 
However, we do not claim \texttt{PDE-READ} needs this many data points to work. 
In particular, \texttt{PDE-READ} works quite well even if $N_{Data}$ is far less than $10,000$. 
For example, in section \ref{sub_sec:Burgers}, we demonstrate that \texttt{PDE-READ} discovers Burgers' equation with $100$\% noise using just $4,000$ data points. 
Further, \texttt{PDE-READ} discovers Burgers' equation with up to $40$\% noise using just $1,000$ data points. 

$~$

\noindent\textbf{Computational Costs of Collocation Points:} In section \ref{sub_sec:Method_Discuss}, we state that $N_{Coll}$ should be as large as is practical. 
In our experiments, $N_{Coll}$ dominates the run-time of \texttt{PDE-READ}. 
Further, the computational cost of enforcing the collocation condition rises with the order of the learned PDE. 
These results are a direct consequence of the fact that \texttt{PDE-READ} must evaluate $U$, along with its spatial partial derivatives, at each collocation point. 
By contrast, at each data point, \texttt{PDE-READ} must only evaluate $U$. 
Differentiating $U$ is expensive. 
Thus, evaluating $N$ at a collocation point is far more expensive than evaluating $U$ at a data point, which explains why $N_{Coll}$ dominates run-time. 
The order of the learned PDE exacerbates this problem, since higher-order equations require higher-order derivatives. 
As such, the computational cost associated with each collocation point increases with the order of the learned PDE. 
Because of these fundamental issues, there is a practical limit to how large $N_{Coll}$ can become before the run-time is unacceptably large.

$~$

Notwithstanding, we can hardly understate the importance of using as many collocation points as possible. 
Increasing $N_{Coll}$ forces $N$ to learn the hidden PDE, and incentivizes $U$ to learn a function that satisfies both the learned PDE and the measurements. 

$~$

Selecting a good value of $N_{Coll}$ requires balancing computational demands with approximation quality.
We ran all of our experiments on one of three computers: 
\begin{enumerate}
    \item A 2015 MacBook Pro with a dual core Intel(R) Core(TM) i7-5557U CPU @ 3.10 GHz CPU. This computer has an integrated GPU and does not support CUDA. As such, all experiments on this computer run on the CPU. All Burgers' and heat equation experiments ran on this machine. 
    \item A high-memory E2 GCP instance. This machine features a dual core CPU (with similar specs to the one in the MacBook Pro) but no GPU. All KdV equation experiments ran on this machine.
    \item A high-performance desktop from Lambda Labs with a 32-core AMD Ryzen Threadripper PRO 3975WX CPU and four Nvidia RTX A4000 graphics cards. All experiments on this machine use GPUs, and run more than ten times faster than experiments on the other machines. All Allen-Cahn, Klein-Gordon, and dynamic beam equation experiments ran on this machine. 
\end{enumerate}

The aging hardware on the first two computers, which ran many of our experiments, limits us to just a few thousand collocation points in most experiments.
This limitation highlights the benefits of collocation point re-selection, which effectively adds additional collocation points without increasing run-time. 
However, increasing $N_{Coll}$ may enable \texttt{PDE-READ} to learn the PDEs in section \ref{sec:Results} at higher noise and sparsity levels than we report.

$~$

\noindent\textbf{\texttt{PDE-READ} Works if $N$ Includes ``Too Many'' Derivatives:} In our experiments, the number of spatial partial derivatives of $U$ that $N$ takes as input matches the order of the hidden PDE.
If the hidden PDE is of order $n$, then $N$ is a function of $U$, and its first $n$ spatial partial derivatives. 
In practice, however, the order of the hidden PDE may be unknown. 
To be practical, \texttt{PDE-READ} must work in this case: which it does.

$~$

To demonstrate this, we use Burgers' equation (a second-order equation).
In these experiments, $N$ accepts $U$ and its first three spatial partial derivatives. 
Further, the library of candidate PDE terms is 

$$\left\{ \left(U\right)^{p_0}\left(D_x U\right)^{p_1}\left(D_x^2 U\right)^{p_2}\left( D_{x}^3 U \right)^{p_3} : p_0, p_1, p_2, p_3 \in \mathbb{N}_{0} \text{ and } p_0 + p_1 + p_2 + p_3 \leq 2 \right\}.$$

First, we corrupt the first Burgers' equation data set with $30$\% noise and then randomly select $10,000$ points from the noisy data set ($N_{Data} = 10,000$). 
We set $N_{Select} = 1$ and $N_{Coll} = 10,000$. 
We train for $2,000$ epochs using the \texttt{Adam} optimizer, with a learning rate of $0.001$, followed by $100$ epochs using the \texttt{LBFGS} optimizer, with a learning rate of $0.1$. 
After training, \texttt{PDE-READ} uses its sparse regression algorithm to identify candidate PDEs.
The highest-ranking candidate PDE is 

$$D_{t} U = (0.078569)(D_{x}^2 U) - (0.851418)(U)( D_{x} U),$$

which matches our form of Burgers' equation (equation \ref{eq:Burgers}).
In this experiment, $N$ learns to ignore the third derivative of $U$. 
With that said, \texttt{PDE-READ} appears to be less robust to noise in this case. 
If we repeat this experiment with $50$\% noise, then the highest-ranking candidate is

$$D_{t} U = -(0.843747)(U)(D_{x} U) + (0.011080)(U)(D_{x}^3 U) - (0.033376)(D_{x} U)(D_{x}^2 U),$$

which does not match our form of Burgers' equation. 
Thus, \texttt{PDE-READ} appears to work best when the number of spatial partial derivatives of $U$ that $N$ takes as input matches the order of the hidden PDE.

$~$

In our tests, if the number of spatial partial derivatives of $U$ that $N$ accepts is less than the order of the hidden PDE, then \texttt{PDE-READ} fails to conclusively identify any PDE. 
In this case, the value of the ranking metric, equation \ref{eq:Ranking_Metric}, for the identified PDE is low ($< 20$\%) and the least-squares residual is large.
In other words, no clear candidate emerges. 

$~$

This suggests the following strategy: First, guess a lower bound for the order of the hidden PDE. 
Set $N$ to accept that many spatial partial derivatives of $U$, and then run \texttt{PDE-READ}. 
If the value of the ranking metric, equation \ref{eq:Ranking_Metric}, for the identified PDE is low, repeat the training process but allow $N$ to accept one more spatial partial derivative of $U$. 
\texttt{PDE-READ} will conclusively identify the hidden PDE once the number of spatial partial derivatives of $U$ that $N$ accepts matches that of the hidden PDE. 

$~$

\noindent\textbf{Network Architecture:} We use the same network architecture in all our experiments. 
In particular, $U$ has five hidden layers with $50$ hidden units per layer, while $N$ has two hidden layers with $100$ hidden units per layer. 
We do not claim this architecture is optimal. 
We originally chose this architecture because Raissi \cite{raissi2018deep} uses it in his two-network approach. 
We retained it because it seemed effective during our numerical experiments. 
More complicated architectures may be necessary when the system response function is highly intricate, while simpler architectures may be beneficial when the system response function is simple.

$~$

{\textbf{The Shrinking Coefficient Problem:}} In almost all of our experiments, the coefficients in the identified PDE are smaller than those in the true PDE.
Further, in most experiments, the shrinkage is roughly the same for each coefficient. 
In other words, if one coefficient in the identified PDE is roughly $10$\% smaller than the corresponding coefficient in the true PDE, then the same is true of the other coefficients in the identified PDE.
Our experiments in section \ref{sec:Results} also suggest that this effect becomes more significant as the noise increases. 
We call this phenomena \emph{shrinking coefficient problem}.

$~$

{As an example, consider our experiments with the Allen-Cahn equation: The} true PDE is 

$$D_t U = 0.003(D_x^2 U) + U - (U)^3,$$

{while the highest-ranking candidates in our $50$\% and $100$\% noise experiments are}

$$D_t U = 0.002040(D_x^2 U) + 0.690463(U) - 0.652140(U)^3$$

and 

$$D_t U = 0.001015(D_x^2 U) + 0.281007(U) - 0.211380(U)^3,$$

respectively. 

$~$

We believe the shrinking coefficient problem occurs because \texttt{PDE-READ} tends to mollify large time derivatives.
If we look at figure \ref{fig:Allen_Cahn:100}, we see that the biggest errors between the learned solution, $U$, and the system response function, $u$, occur in regions where $D_t u$ is large.
Indeed, a close inspection of figures \ref{fig:Allen_Cahn_Dataset} and \ref{fig:Allen_Cahn:100} reveal that $U$ takes longer to transition from blue to red, and vice versa, than $u$.
This naturally raises two questions: Why does this cause the coefficients in the identified PDE to shrink, and why does it occur in the first place?
To answer the first question, consider the form of the hidden PDE,

\begin{equation} D_{t}^n u = \hat{N}(u, \ldots, D_{x}^{M} u ) \label{eq:PDE:Discuss} \end{equation}

If $D_{t} U$ is, on average, smaller than $D_{t} u$, then we expect the output of $N$ to be smaller, on average, than that of $\hat{N}$.
Since we determine the coefficients in the identified PDE by fitting $N$ to the library of candidate PDE terms, {the existence of the shrinking coefficient problem suggests that $D_{t} U$ (and, therefore, $N$) is, on average, smaller than $D_{t} u$.
Interestingly, the shrinking coefficient problem seems to be limited to time derivatives. In particular, our experiments with Burgers' equation demonstrate that \texttt{PDE-READ} can learn sharp spatial derivatives without mollifying them.}

$~$

{We are not entirely sure what causes $D_t U$ to be smaller than $D_t u$. 
We, however, suspect the problem stems from the noise and the collocation loss.
The collocation loss} effectively enforces equation \ref{eq:PDE:Discuss}.
If $D_{t} U$ takes on large values, then $N$ must also take on large values.
{In our data sets, large} time derivatives are relatively rare. 
Across most of the data set, the $D_t u$ is relatively small.
In other words, a large $D_t u$ value is somewhat of an outlier, and $U$ appears to be treating it as such; rather than fitting large $D_t U$ values, {our plots} suggest that $U$ attempts to ignore them.
This causes $D_t U$ to be smaller than $D_t u$ in the few regions where the latter is large.
If $D_t U$ is smaller than $D_t u$, then $U$ will fail to change as quickly (with respect to time) as $u$, which is precisely what we see with the Allen-Cahn equation. 
Thus, seeking to minimize the data loss pressures $D_t U$ to match $D_t u$, even in regions where the latter is large.
These two factors compete with one another, though the former appears to be strong enough to shrink the coefficients in the identified PDE.

$~$ 

{Our hypothesis may also explain why noise exacerbates the shrinking coefficient problem.
At low noise levels, the solution network effectively learns the data exactly. 
At higher noise levels, however, the solution network effectively learns the average of the noise, rather than the data set itself.
This suggests that as the noise level increases, the solution network is influenced less by the solution and more by the noise signal within the data.}
This means the data loss puts less pressure on $D_t U$ to match $D_t u$.
Thus, noise effectively tips the scale in favor of $U$ treating large time derivatives as outliers, thereby resulting in even smaller coefficients in the identified PDE. 

$~$

{Finally, it is worth noting that the shrinking coefficient problem is far less significant in low-noise settings.
To demonstrate this, we run two additional experiments, one for the Allen-Cahn equation and the other for Burgers' equation.
For the Burgers' equation experiment, we use the first Burgers' equation data set, from Section 4. 
In both of the new experiments, we use  $10,000$ data points with $0$\% noise and $5,000$ collocation points.
Further, in these experiments, we train the networks for $2,000$ epochs using the \texttt{Adam} optimizer with a learning rate of $0.001$, followed by $400$ epochs of the \texttt{LBFGS} optimizer with a learning rate of $0.1$.
Finally, in these experiments, our library of candidate terms is}
$$\left\{ \left(U\right)^{p_0}\left(D_x U\right)^{p_1}\left(D_x^2 U\right)^{p_2}\left(D_x^3 U\right)^{p_3}\left(D_x^4 U\right)^{p_4} : p_0, p_1, p_2, p_3, p_4 \in \mathbb{N}_{0} \text{ and } p_0 + p_1 + p_2 + p_3 + p_4 \leq 2 \right\}.$$
{After training, the highest-ranking candidate PDE in the Allen-Cahn experiment is}
$$D_t U = 0.968938(U) + 0.002924(D_x^2 U) -0.961979(U)(U)(U),$$
{while the highest-ranking candidate in the Burgers' equation experiment is}
$$D_t U = 0.095125(D_x^2 U) -0.965024(U)(D_x U).$$
{In both cases, the coefficients in the identified PDE are roughly $4$\% smaller than the corresponding coefficients in the true PDE.
Thus, the shrinking coefficient problem persists in the low-noise limit but is certainly less significant.}

$~$

While {the shrinking coefficient problem} is not desirable, it does not appear to be detrimental either. 
In particular, our experiments demonstrate that \texttt{PDE-READ} consistently identifies the correct PDE, even if the coefficients in the identified PDE are considerably smaller than those in the true PDE. 
In other words, the shrinking coefficient problem does not appear to impair \texttt{PDE-READ}'s ability to fulfill its primary objective: to identify the form of the hidden PDE.
With that said, the shrinking coefficient problem is certainly not desirable{.
Determining its root cause and developing strategies to mitigate it represents an important area of future research. }

\subsubsection{Heat Equation} 
\label{sub_sub_sec:Heat_Discuss}

In section \ref{sub_sec:Heat}, the highest-ranking candidate after training is 

\begin{equation} D_t U = -(0.459)(U), \label{eq:Wrong_Heat} \end{equation}

while the true PDE (the one we generate the data set from) is the following:
$$D_t U = -(.05)(D_{x}^2 U).$$
Thus, it appears that \texttt{PDE-READ} fails, but the truth is more interesting, and reveals a fundamental limitation of PDE discovery. 

$~$

To begin, our sparse regression algorithm is extremely confident in the identified PDE.
According to the ranking metric (see equation \ref{eq:Ranking_Metric}), removing the LIF from the highest-ranking PDE causes its least-squares residual to increase by more than $7,000$\%.
By contrast, removing the LIF from the second highest-ranking PDE increases its residual by just a few percent. 
Thus, the identified PDE fits $N$ exceptionally well.
We can understand what is happening if we solve the heat equation using the method of \emph{separation of variables}.
We can express the solution, $U$, as 

$$U(t, x) = T(t)X(x).$$

Substituting this into the heat equation and rearranging gives the following:

$$\frac{T'(t)}{T(t)} = c\frac{X''(x)}{X(x)}$$
$$X(0)T(t) = X(10)T(t)$$
$$X'(0)T(t) = X'(10)T(t).$$

The only way this can be true for all $t, x$ is if both sides equal a constant, $\lambda$. 
This forces $X$ and $T$ to satisfy the following ODEs:

$$T'(t) = \lambda T(t)$$
\begin{equation}X''(x) = \left(\frac{\lambda}{c}\right) X(x) \label{eq:DDX} \end{equation}

Further, by virtue of the initial conditions and continuity, $T$ must be non-zero.
Thus,

\begin{equation} X(0) = X(10) \label{eq:XBC} \end{equation}
$$X'(0) = X'(10).$$

If we substitute these relationships back into the original PDE, we find that 

\begin{align*} \frac{\partial U(x, t)}{\partial t} &= X(x)T'(t) \\
&= c T(t)X''(x) \\ 
&= \lambda T(t) X(t) \\
&= \lambda U(x, t). \end{align*}

Further, the initial condition becomes the following:

$$U(0, x) = T(0)X(x) = \sin\left(\frac{(x - 5) 2 \pi}{10} \right).$$

If we solve equation \ref{eq:DDX} subject to equation \ref{eq:XBC} and impose the above initial conditions, we find that the only suitable value of $\lambda$ is $\pi^2 c$, which is approximately $0.493$. 
Thus, the system response function in this data set satisfies the following PDE:

$$\frac{\partial U(t, x)}{\partial t} = (\pi^2 c )U(t, x)$$

which is almost exactly the identified PDE. 

$~$

This {highlights a fundamental aspect of PDE-Discovery}: The system response function, $u$, may satisfy multiple PDEs. 
As an especially pathological example, consider the exponential function. 
We know $\exp$ satisfies 

$$U'(x) = U(x),\quad\quad U(0) = 1,$$

as well as

$$U''(x) = U(x),\quad\quad U(0) = U'(0) = 1,$$
$$U'''(x) = U(x),\quad\quad U(0) = U'(0) = U''(0) = 1,$$

and even 

$$\frac{1}{3}(U' + U'' + U''') = U,\quad\quad U(0) = U'(0) = U''(0) = 0.$$

In fact, any linear combination of the first three IVPs yields another IVP whose solution is $\exp$. 
Thus, $\exp$ satisfies an unaccountable infinity of IVPs. 
Trying to identify \emph{the} ODE which describes the evolution of $\exp$ is a fundamentally ill-posed problem.

$~$

{While we explicitly constructed this example and the first heat equation data set to highlight this issue, they nonetheless demonstrate that non-uniqueness can arise in PDE-discovery applications.
This seeming limitation does not mean that PDE discovery a fundamentally flawed task, however.
Rather, it highlights the fact that, as with many inverse problems, multiple solutions may exist.}

$~$

{It is worth noting that in the case of the first heat equation data set,} \texttt{PDE-READ} learns a simpler (in the sense that it is lower order) PDE. 
This strongly suggests \texttt{PDE-READ} seeks the simplest PDE whose solution is the system response function; \texttt{PDE-READ} seems to be biased towards parsimonious PDEs. 
This makes sense. 
The identified PDE is first order, while the heat equation itself is second-order. 
To learn the identified PDE, U only needs to learn to match the system response function. 
By contrast, to learn the heat equation, $U$ has to learn the system response function, along with its first two spatial partial derivatives: a more challenging feat. 
Our results are consistent with \cite{rudy2017data}'s argument that PDE discovery algorithms should identify parsimonious descriptions of the hidden dynamics.

$~$

Notably, \texttt{PDE-READ} identifies the heat equation in the second data set. 
As far as the authors are aware, the heat equation is the simplest PDE that the system response function for this data set satisfies.

\subsubsection{Burgers' Equation} 
\label{sub_sub_sec:Burgers_Discuss}

Our experiments with Burgers' equation demonstrate that \texttt{PDE-READ} is robust to sparsity and noise. 
As far as the authors are aware, no existing PDE discovery tool can identify Burgers' equation with $100$\% noise using just $4,000$ data points. 
The closest example we are aware of is DeepMoD \cite{both2021deepmod}.
DeepMoD demonstrates impressive robustness to sparsity and noise. 
However, when training on just $4,000$ data points, DeepMoD struggles to identify Burgers' equation with more than $50$\% noise. 

$~$

Our experiments with the first Burgers' equation data set suggest that \texttt{PDE-READ}'s robustness to noise decreases as sparsity increases. 
This makes sense. 
Identifying the system response function is difficult if there are a limited number of measurements. 
Noise only exacerbates this challenge. 
Recall that we add noise by sampling a zero-mean Gaussian distribution.
If the density of data points is high enough, then the best way to minimize the data loss is for $U$ to learn the {\it average} of the noise, which is the system response function. 
By contrast, if the density of data points is too small, then separating the system response function from the noise is difficult.

$~$

\begin{figure}[ht!]
    \centering
    \includegraphics[width=\linewidth]{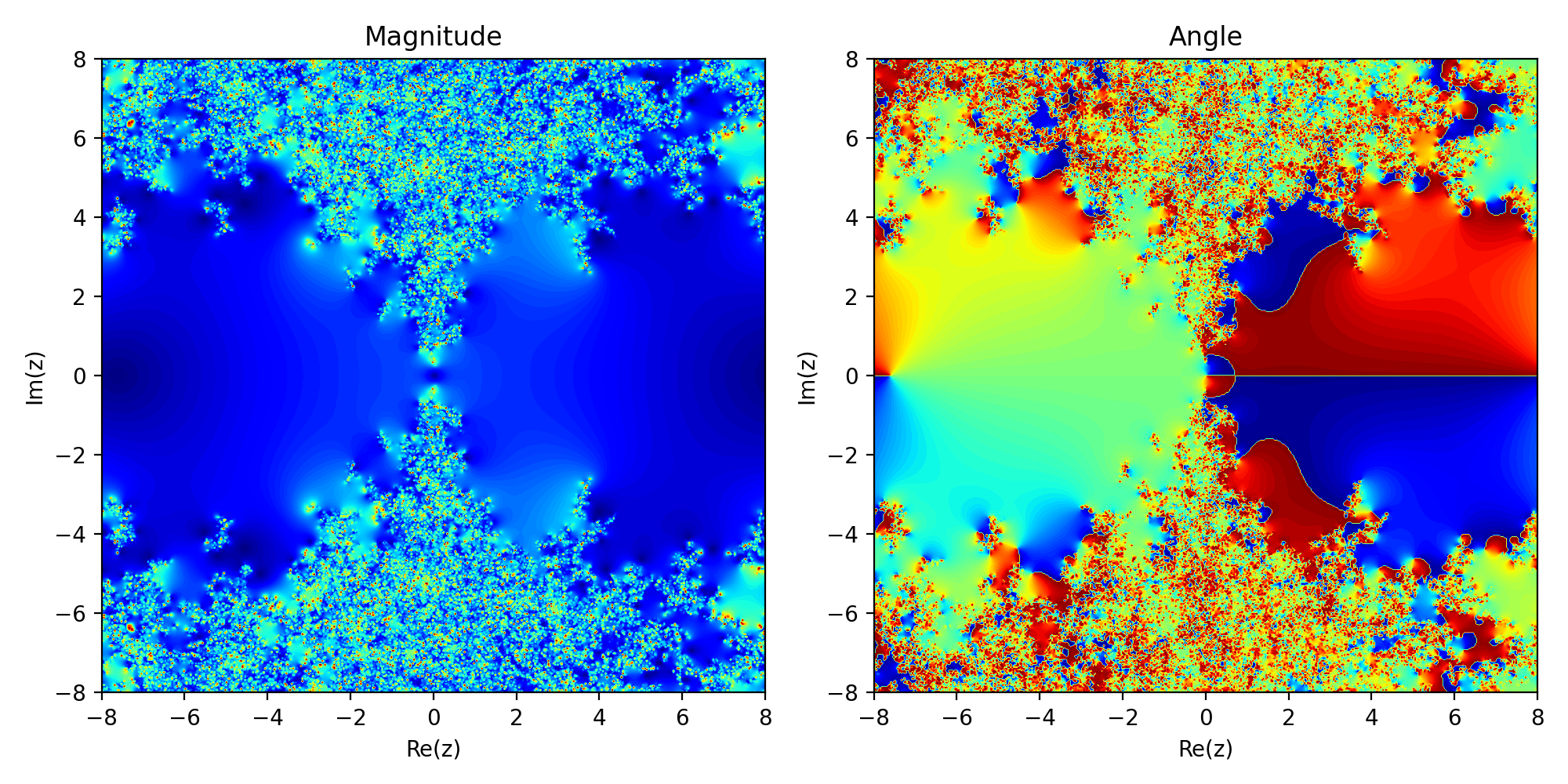}
    \caption{These plots depict the modulus and argument of the range of the function $x \to U(4.0, x)$, for $x \in \mathbb{C}$. Here, $U$ is the solution network after training on the first Burgers' data set with $100$\% noise, and $10,000$ measurements.}
    \label{fig:Rational_Plot:Burgers}
\end{figure}

Additionally, \texttt{PDE-READ}'s unique RatNN architecture affords additional physical insight; the first Burger's equation data set contains a shock at $x = 0$, which forms around $t = 3$.
An interesting property of RatNNs is that their poles can reveal information about discontinuities in the network and by extension, the system response.
\cite{boulle2022data} discusses this in detail.
Consider a RatNN, $U$, with two input variables, $x$ and $t$.
Fix $t = t_0$ for some $t_0 \in \mathbb{R}$. 
The function $x \to U(t_0, x)$ for $x \in \mathbb{C}$ is rational since the composition of a sequence of rational functions is rational.
As the network trains, the poles of this function move. 
Their locations can inform us about the discontinuities in the learned solution at $t_0$.
To demonstrate this, we fix $t_0 = 4.0$ and consider the function $x \to U(4.0, x)$ for $x \in \mathbb{C}$ (here, $U$ is the network from the $100$\% noise experiment with the first Burgers equation data set).
Figure \ref{fig:Rational_Plot:Burgers} depicts the magnitude and angle of this function on the domain $[-8, 8]x[-8, 8] \subseteq \mathbb{C}$.

$~$

The only place where the poles approach the real axis is at the origin, reflecting the fact that the system response function contains a discontinuity at that location: the position of our shock front.
In principle, we can use these plots to uncover properties of $u$ and the hidden PDE.
\cite{boulle2022data} discusses this extensively. 

$~$

\subsubsection{Allen-Cahn Equation}
\label{sub_sub_sec:AC_Discuss}

Our experiments with the Allen-Cahn equation are significant because the Allen-Cahn equation is the only equation we study that contains a third degree term (namely, $U^3$).
In the other non-linear equations we considered, the non-linearity can be expressed as the product of two elements of $\{ U, D_x U, \ldots, D_x^M U \}$. 
Thus, in a sense, the Allen-Cahn equation is the most non-linear equation we consider.
Nonetheless, \texttt{PDE-READ} is able to identify the Allen-Cahn equation, even with $100$\% noise. 

$~$

\subsubsection{Korteweg-De Vries Equation} 
\label{sub_sub_sec:KdV_Discuss} 

Our experiments with the KdV equation are significant because they demonstrate that our sparse regression algorithm can identify the correct PDE from a {\it very} large library of candidate PDE terms. 
In our $10$\% and $60$\% experiments with the KdV equation, our library contained $126$ terms. 
The theorem in section \ref{sub_sub_sec:Sparse_Discuss} tells us that our sparse regression algorithm always removes the feature that is least important in minimizing the least-squares residual.
Our experiments with the KdV equation suggest that this property translates into a robust, practically useful algorithm.

$~$

Moreover, the $75$\% and $100$\% experiments demonstrate \texttt{PDE-READ}'s unprecedented ability to learn from noisy data sets.
The KdV data set contains subtle features, many of which are practically imperceptible to the human eye in the noisy data set (see the upper left sub-plot of figure \ref{fig:KdV:100}). 
Nonetheless, \texttt{PDE-READ} manages to uncover these subtle features and identify the KdV equation. 

$~$

\subsubsection{Klein-Gordon Equation}
\label{sub_sub_sec:KG_Discuss}

Our experiments with the Klein-Gordon equation (and the dynamic beam equation) is significant in that they demonstrate \texttt{PDE-READ} can learn PDEs with higher-order time derivatives. 
Further, the fact that we can identify the KG equation with up to $100$\% noise demonstrates that \texttt{PDE-READ} does not lose its robustness to sparsity and noise when dealing with higher-order time derivatives.

$~$

Significantly, the coefficients in the identified PDE are very similar to those of the true PDE, even in the $100$\% noise experiment. 
Unlike the five other equations we consider, \texttt{PDE-READ} does not seem to suffer from the small coefficient problem while learning the KG equation. 
There are a few possible reasons for this.
Most significantly, $N$ is learning an acceleration (second time derivative) in this case, rather than a velocity (first time derivative).
The second time derivative of $u$ may be large over a big enough portion of $\Omega$ that $N$ can not afford to treat large accelerations as outliers.
Whatever the reason, this result suggests that \texttt{PDE-READ} can learn accurate coefficients, even in the face of extreme noise (though as our experiments with the other equations show, this is generally not the case). 

$~$

\subsubsection{Dynamic Beam Equation}
\label{sub_sub_sec:Beam_Discuss}

Finally, our experiments with the dynamic beam equation are significant because the dynamic beam equation is the highest-order equation we tested. 
These experiments indicate that \texttt{PDE-READ} can learn accurate fourth order spatial partial derivatives, along with second-order temporal derivatives, from very noisy data sets.
This is important as derivatives tend to amplify noise, making it harder to separate the underlying signal from that noise.
Notwithstanding, \texttt{PDE-READ} successfully identifies the dynamic beam equation at $50$\% noise. 
While it is worth noting that \texttt{PDE-READ} appears to be less robust to noise in the dynamic beam equation data set, as compared with the other data sets we considered (it is the only equation which \texttt{PDE-READ} could not reliably identify at $100$\% noise), the very fact that it can learn a fourth order equation from noisy data is significant.

\section{Conclusion} 
In this paper, we have introduced \texttt{PDE-READ}, a novel algorithm that uses a principled, parameter-free sparse regression algorithm to identify human-readable PDEs from data. 
\texttt{PDE-READ} achieves this end by introducing a variant of Raissi's two-network approach \cite{raissi2018deep} (that employs RatNNs and resampling of random collocation points within the problem domain) and a modified version of the Recursive Feature Elimination algorithm \cite{guyon2002gene}.
Our numerical experiments demonstrate that \texttt{PDE-READ} can identify a diverse collection of linear and non-linear PDEs, even when faced with significant sparsity and noise.
Our approach also demonstrates the efficacy of RatNNs and collocation point re-selection in PDE discovery. 

$~$

Additionally, our sparse regression algorithm is parameter-free: there are no parameters to fine-tune. 
We believe this gives our algorithm a distinct advantage over existing sparse regression algorithms, such as STRidge and LASSO. 
Our algorithm's favorable theoretical properties further strengthen this belief.

$~$

Based on the foregoing observations, we believe \texttt{PDE-READ} may be useful in discovering predictive models of complex physical systems that currently lack predictive models. 

\section{Acknowledgements}
This work is supported by the Office of Naval Research (ONR), under the grants N00014-19-1-2034 and N00014-22-1-2055.

\printbibliography 

\end{document}